\documentclass[10pt,twocolumn,letterpaper]{article}

\usepackage[pagenumbers]{cvpr} % arXiv version (de-anonymized, page numbers)

\usepackage[utf8]{inputenc}
\usepackage[T1]{fontenc}
\usepackage{graphicx}
\usepackage{amsmath}
\usepackage{amsfonts}
\usepackage{booktabs}
\usepackage{array}
\usepackage{nicefrac}
\usepackage{microtype}
\usepackage{xcolor}
\usepackage{subcaption}
\usepackage{enumitem}
\usepackage{multirow}
\usepackage{url}

\definecolor{cvprblue}{rgb}{0.21,0.49,0.74}
\usepackage[pagebackref,breaklinks,colorlinks,allcolors=cvprblue]{hyperref}

\def\confName{CVPR}
\def\confYear{2027}

\usepackage{pgfplots}
\pgfplotsset{compat=1.17}
\begin{document}
\title{No Safe Dose: How Training Data Drives Unsafe Image Generation}
\author{
Felix Friedrich$^{1,2}$ \quad
Lukas Helff$^{2}$ \quad
Niharika Hegde$^{2,3}$ \\
Patrick Schramowski$^{2,3,4}$ \quad
Kristian Kersting$^{2,3,4}$ \\[4pt]
{\normalsize $^{1}$Black Forest Labs \quad $^{2}$TU Darmstadt \& hessian.AI \quad $^{3}$DFKI \quad $^{4}$Lab1141}
}
\maketitle

\begin{abstract}
Text-to-image models are trained on web-scale data that inevitably contains unsafe images, yet whether this contamination directly drives unsafe generation has not been tested under controlled conditions. We train the same text-to-image model on corpora that differ only in their fraction of unsafe images (0\% to 9.6\%) and in size (100K to 8M), and evaluate 10,000 generations per model with four independent safety classifiers. Output unsafety rises monotonically and saturates, from 16.6\% at 0\% to 26.4\% at 9.6\% contamination, and the effect appears only under adversarial prompts. At a fixed training budget, the level is set by the \emph{proportion} of unsafe images rather than their absolute count or cumulative exposure, so filtering a fixed fraction is equally effective at 1M and 8M images, at no measurable cost in FID, CLIPScore, or ImageReward. Filtering alone, however, does not eliminate unsafe generation: a floor persists when every image flagged by any of four judges is removed (13.6\%) and rises when training on real instead of synthetic images (23.1\%). A safety-aligned text encoder lowers the floor to 9.6\%, but it is not carried by a removable encoder direction, which implicates the generative backbone. Finally, fine-tuning on entirely safe data erodes the benefit of curation ($+8.7$pp). Data curation and encoder choice are complementary and effective, but neither makes a model safe on its own.
% [previous abstract]
% Text-to-image models trained on large-scale data often inevitably ingest unsafe content. It remains unclear whether the unsafe fraction of the training data directly drives output safety, or whether other factors dominate. We shed light on this question by isolating this variable: we train the same text-to-image model on datasets that differ
% \emph{only} in their fraction of unsafe images (0\% to 9.6\%), across several dataset scales (100K to 8M).
% Then we generate images with the resulting models, and evaluate them with four independent safety classifiers. Output unsafety rises monotonically from 16.6\% at 0\% contamination to 25.5\% at 5\%. A factorial design reveals that the \emph{proportion}, not the absolute count, of unsafe training images is the operative variable. The 16.6\% irreducible baseline at zero contamination implicates the other components, e.g. frozen text encoder, as a residual safety risk---confirmed by a text encoder ablation showing that SafeCLIP reduces this floor to 9.6\%, while the dose-response effect persists across all three encoders tested. Critically, no quality degradation in terms of FID, CLIPscore and ImageReward accompanies safety filtering. These results establish that data curation and text encoder safety are complementary and independently effective interventions. At the same time, the remaining level of unsafety poses questions for future research about emerging capabilities and compositionality.
% %\ps{"data curation specifically addresses adversarial robustness" insight is missing}
\end{abstract}

\section{Introduction}
\begin{figure}[t]
\centering
     
    \includegraphics[width=\linewidth]{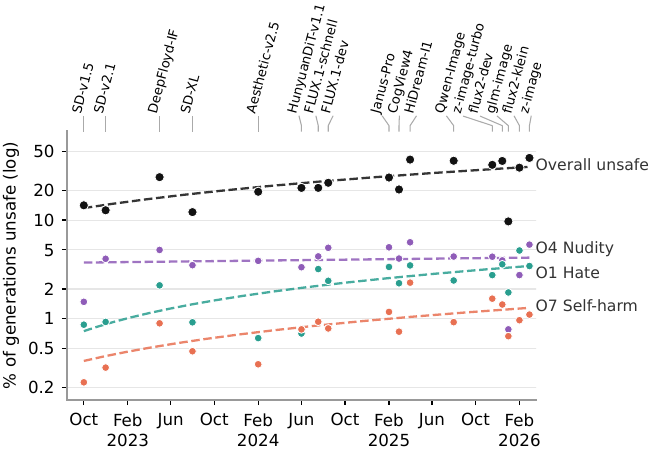}
    \caption{\textbf{Models become less safe over time.} Share of generations judged unsafe by LlavaGuard on our prompt testbench for 17 open-weight T2I models, by release date (log scale); categories are shares of all generations. Dashed: least-squares linear fits. The overall rate rises by ${\approx}6$pp per year (Spearman $\rho{=}0.58$, $p{=}0.015$), with hate and self-harm rising most steeply.}
    \label{fig:observational_motivation}
     
\end{figure}
Text-to-image (T2I) generative models are now deployed at unprecedented scale, with systems such as Nano Banana \cite{nano_banana_2025}, Midjourney \cite{midjourney_2025}, and open-weight models like Flux~\citep{esser2024scaling}, Stable Diffusion~\cite{rombach2022high}, or Qwen-Image \cite{wu2025qwenimagetechnicalreport} serving billions of users. These models are trained on massive, sometimes uncurated large-scale datasets that often inevitably contain unsafe content---images depicting violence, sexual material, hate imagery, and other categories of harm~\citep{birhane2021large,birhane2021multimodal,birhane2024into,schuhmann2022laion}. Concerns about the safety of generated outputs have spurred a range of reactive interventions, including prompt filtering, RLHF, guidance mechanisms, etc.~\citep{schramowski2023safe,poppi2024safe,friedrich2023fair,helff2024llavaguard}. Yet a fundamental upstream question remains open: \emph{to what extent does training data composition directly drive output safety?} The answer governs how data curation should be prioritized relative to other interventions: as a primary lever, a complementary layer, or a marginal contribution, based on its effect-cost tradeoff.

Fig.~\ref{fig:observational_motivation} documents model output safety across 17 T2I models released between October 2022 and early 2026. As one can see, the prevailing "bigger is better" scaling paradigm has largely ignored the safety-quality trade-off inherent in data curation. Instead, we observe a troubling trend: newer, more capable models generate unsafe content at higher rates than predecessors. This puts safety mitigations, e.g.\ dataset curation, even more prominently than ever. However, it is unclear what drives this trend. A direct link between training data and model outputs is so far unavailable and mostly precluded by two main confounding variables: proprietary datasets and varying setups. 

% This works addresses this gap by conduct
% If data composition is a first-order driver of output safety, curation becomes a mandatory intervention; this will also inform current regulatory focus on data auditing.

We address this gap with a controlled experiment that isolates the unsafe fraction of the training data as the variable of interest. We train a fixed text-to-image flow-matching model (PRX) under seven experimental conditions that vary the proportion of unsafe images in the training corpus (0\% to 9.6\%) and the corpus size (100K to 8M), holding everything else constant, e.g.\ architecture, hyperparameters, evaluation protocol, etc. A text encoder ablation (T5-Gemma, CLIP, SafeCLIP) measures how much the conditioning model contributes to the unsafety that remains at 0\% contamination. We generate 10,000 images with each resulting model from our prompt testbench, and outputs are evaluated by four independent safety classifiers.

Our experiments reveal a chain of increasingly specific insights. \textbf{First}, training data contamination directly and monotonically drives unsafe image generation: the more unsafe content in the training set, the more unsafe the outputs. \textbf{Second}, a factorial design shows that, at a fixed training budget, the operative variable is the \emph{proportion} of unsafe images, not their absolute count, and a matched-exposure arm shows that doubling cumulative exposure at fixed proportion does not raise it. In practice, removing a fixed fraction of unsafe content was equally effective at 1M and 8M images. \textbf{Third}, and perhaps most strikingly, the dose-response effect is exclusively adversarial: under safe prompts, all models produce approx.\ 1\% unsafe outputs regardless of contamination level. Contamination is invisible to benign users; it shapes only the adversarial attack surface. \textbf{Finally}, safety filtering incurs no measurable quality cost in FID, CLIPscore, or ImageReward.

Removing every unsafe image, however, does \emph{not} remove unsafe generation: a 16.6\% floor persists at 0\% contamination, and we show that this floor is neither an artifact of how ``unsafe'' was labeled nor a property of our particular data source. Intersecting four independent safety judges---discarding every image that \emph{any} judge flags, 14.7\% of the pool---lowers the floor only to 13.6\%, and just 1.0\% of the admitted pool is called unsafe by a majority of the held-out judges, far too little to account for a 13--23\% output rate. Retraining the 0\% arm on 1.5M \emph{real} web images raises the floor to 23.1\% rather than removing it. The frozen text encoder contributes to, but does not constitute, this floor: replacing T5-Gemma with SafeCLIP lowers it to 9.6\%, yet projecting the learned unsafe direction out of the conditioning leaves the floor intact (15.6\% baseline vs.\ 16.8\% ablated). The residual therefore implicates the generative backbone and its compositional capacity rather than a single localizable channel. Safety is fragile downstream as well: supervised fine-tuning on a small, entirely safe, real-image corpus erodes it by $+8.7$pp for the 0\% model, the largest increase of any condition at that scale.

Summarized, our key contributions are\footnote{We publicly release all trained models, generated images and annotations at {\tiny\url{https://huggingface.co/collections/anonym371/no-safe-dose}}. Access is gated for research purposes, see impact statement.}: 
\textbf{(i)} We provide a controlled analysis of how training data composition and scale affect output safety in T2I models. 
\textbf{(ii)} We demonstrate a direct causal relationship between dataset safety and model output safety. 
\textbf{(iii)} Through extensive ablations, we demonstrate that safety filtering incurs minimal quality trade-offs, that unsafe generation persists under the strictest multi-judge filtering and on real images, and that post-training erodes the benefit of curation.

% \begin{figure}[t]
%     \centering
%     \begin{subfigure}[t]{0.48\linewidth}
%         \centering
%         \includegraphics[width=\linewidth]{figures/timeline_safety_abs_rel_datasets.png}
%         \caption{Datasets: relative unsafe proportion stays flat ($\sim$1--2\%), but absolute unsafe image count grows dramatically with dataset scale.}
%         \label{fig:motivation_datasets}
%     \end{subfigure}
%     \hfill
%     \begin{subfigure}[t]{0.48\linewidth}
%         \centering
%         \includegraphics[width=\linewidth]{figures/timeline_models_rising_cats.png}
%         \caption{Models: unsafe generation rates rise across successive T2I model generations, with certain harm categories showing steeper increases.}
%         \label{fig:motivation_models}
%     \end{subfigure}
%     \caption{\textbf{Observational motivation.} We previously documented two trends across 10 datasets and 12 models spanning 15 years: (a) absolute unsafe content in training data grows with dataset scale, and (b) newer models generate more unsafe content. This paper tests whether the correlation is driven by training data composition through a controlled experiment.}
%     \label{fig:observational_motivation}
% \end{figure}

\section{Related work}

\textbf{Training data quality and its downstream effects.}
Large-scale datasets have been shown to contain substantial amounts of problematic content, including hate speech, pornography, and stereotypical representations~\cite{birhane2021large,birhane2021multimodal,birhane2024into}. The datasheets framework~\citep{gebru2021datasheets} formalized the need for dataset documentation, and automated auditing tools have been proposed to scale this effort~\citep{schramowski2022can}. Beyond documentation, a growing body of work demonstrates that generative models \emph{amplify} distributional properties of their training data: \citet{hall2022systematic} showed systematic bias amplification, and \citet{steed2021image} found human-like biases in unsupervised image representations. Others \citep{luccioni2023stable,friedrich2025divbench,struppek2023exploiting} further document systematic societal biases in T2I diffusion outputs through controlled probes. Scaling laws~\citep{kaplan2020scaling} establish that absolute dataset size drives learning dynamics, while others document and demonstrated that diffusion models can memorize and reproduce individual training images \citep{carlini2023extracting,somepalli2023diffusion,hintersdorf2024clip}. Collectively, this literature supports the data-centric view that training corpus composition matters, but prior work has missed to isolate data contamination fraction as a clear controlled variable.

\textbf{Safety evaluation and mitigation in T2I models.}
Safety classifiers for generated images include vision-language models such as LlavaGuard~\citep{helff2024llavaguard} and LlamaGuard~\citep{inan2023llamaguard}, as well as specialized content moderation models like ShieldGemma~\citep{zeng2025shieldgemma}. Mitigation strategies span multiple layers: Safe Latent Diffusion~\citep{schramowski2023safe,brack2023sega} and LEdits++~\citep{brack2024ledits} intervene during the diffusion/flow process; Safe-CLIP~\citep{poppi2024safe} modifies the embedding space; prompt-level filtering and modification blocks adversarial inputs before generation \citep{friedrich2023revision}, illustrate analogous principles for steering generative behavior. A parallel line of work targets the model weights directly through concept erasing and ablation \citep{gandikota2023erasing,kumari2023ablating,zhang2023forgetmenot,gandikota2024unified,lu2024mace}, which fine-tune pretrained diffusion models to suppress specific unsafe concepts without retraining from scratch. Post-training preference alignment \citep{black2024training,wallace2024diffusiondpo,lee2023aligning,nakamura2025aurora} offers a further complementary avenue, as do recent feature-level guidance methods based on interpretability \citep{harle2025measuring}. Red-teaming efforts have developed systematic methods for discovering failure modes~\citep{rando2022redteaming,li2024art,quaye2024adversarial,qu2023unsafe,yang2024sneakyprompt,struppek2023exploiting,tedeschi2024alert,friedrich2024malert,rottger2025msts}. These approaches are \emph{reactive}, they accept the training data as given and mitigate downstream. We study the complementary \emph{upstream} question of how training data composition shapes model safety prior to those interventions.

% \textbf{Safety benchmarks and evaluation frameworks.}
% A broader literature studies how safety should be measured and governed across generative systems. \citet{solaiman2023evaluating} propose a framework for evaluating the social impact of generative AI in both technical and societal terms; \citet{reuel2025who} survey first- and third-party evaluation coverage and identify systematic gaps; and \citet{mundt2025cake} discuss participation and governance through an extended metaphor for who builds and audits AI systems. On the benchmark side, MLCommons AILuminate~\citep{ghosh2025ailuminate} aggregates risk-and-reliability criteria into a standardized industry benchmark, 

\textbf{Controlled experiments on training data composition.}
Controlled experiments manipulating training data composition for generative models remain rare. \citet{seshadri2023bias} identified the "bias amplification paradox," revealing that apparent increases in bias are often artifacts of distribution shifts between training captions and inference prompts. \citet{Brack_2025_ICCV} applied this observation downstream and studied the effect of altering gender distributions in training captions. Broader audits by \citet{friedrich2023fair} have examined the interplay between dataset bias, compositionality, and generative outputs, demonstrating how ingrained biases in large-scale scraped data propagate to model behavior. The data-centric AI paradigm~\citep{gadre2024datacomp} has motivated systematic ablations of dataset composition for discriminative models \citet{hall2022systematic}. Despite these insights into bias and classifiers, equivalent controlled experiments for safety and generative models are absent.
In contrast, this paper isolates the fraction of unsafe training images as an independent variable for T2I model training.

\section{Method}
\label{sec:method}

We ask whether the fraction of unsafe training images causally affects unsafe image generation in text-to-image models. Our method formalizes (i) the dose variable induced by training-data contamination, (ii) the outcome variable measuring unsafe generations under fixed prompt distributions, and (iii) a controlled intervention family that enables identification of the effect of contamination proportion separately from absolute unsafe count.

Let \(\mathcal{D}=\{(x_i,c_i)\}_{i=1}^{N}\) be an image--text corpus with images \(x_i \in \mathcal{X}\) and captions \(c_i\in\mathcal{C}\). And let $A:\mathcal{X}\to\{0,1\}$ be a binary \emph{intervention labeler} that flags whether an image is unsafe under an operational policy. The \emph{training contamination rate} (``dose'') of \(\mathcal{D}\) is
\begin{equation}
p(\mathcal{D}) = \frac{1}{N}\sum_{i=1}^{N} A(x_i) \in [0,1] \,,
\label{eq:dose}
\end{equation}
and the absolute number of unsafe training examples is
\begin{equation}
U(\mathcal{D}) = \sum_{i=1}^{N} A(x_i) = N\,p(\mathcal{D}) \,,
\label{eq:count}
\end{equation}
which equals the fraction times the dataset size.

Let \(f_{\theta}(\cdot \mid t)\) denote a T2I model trained on \(\mathcal{D}\), producing an image \(y\) conditioned on a prompt \(t\). For evaluation, let $G:\mathcal{X}\to\{0,1\}$ be a (possibly different) \emph{judge} that flags unsafe \emph{generated} images. For a prompt distribution \(\pi\) over text prompts, define the unsafe generation rate
\begin{equation}
q_G(\theta;\pi)
=
\Pr_{t\sim \pi,\; y \sim f_{\theta}(\cdot\mid t)}\big[ G(y)=1 \big] \,.
\label{eq:unsafe_rate}
\end{equation}

\noindent\textbf{Primary estimand (dose--response).}
For each judge \(G\) and prompt distribution \(\pi\), we seek functional relationship
\begin{equation}
q_G(p;\pi)= \mathbb{E}\big[q_G(\theta(p);\pi)\big]\,,
\label{eq:dose_response}
\end{equation}
where \(\theta(p)\) denotes parameters obtained by training on a dataset whose contamination rate is \(p\), and the expectation ranges over training stochasticity and sampling randomness.

\textbf{Secondary estimand (proportion vs.\ count).}
Because \(U=N\,p\), observational comparisons typically confound ``fraction unsafe'' with ``more unsafe images in total.'' Our goal is to isolate whether unsafe generation is principally driven by the proportion \(p\) rather than the count \(U\), while accounting for scale \(N\).

\subsection{Intervention family: controlled contamination of the training distribution}
\label{sec:method:intervention}

Starting from a fixed base corpus \(\mathcal{D}_{\mathrm{base}}\), we apply intervention labeler \(A\) to partition examples into un/safe subsets:
\[
\begin{aligned}
\mathcal{S} &= \{(x,c)\in \mathcal{D}_{\mathrm{base}}: A(x)=0\}\,,\\
\mathcal{U} &= \{(x,c)\in \mathcal{D}_{\mathrm{base}}: A(x)=1\}\,.
\end{aligned}
\]
We then define a family of datasets \(\{\mathcal{D}(p,N)\}\) by adjusting the mixture weight of \(\mathcal{U}\) relative to \(\mathcal{S}\) to attain a target contamination rate \(p\) at a specified scale \(N\). Conceptually, this constitutes an intervention on the training distribution that changes the prevalence of unsafe visual concepts while preserving all other aspects of the training pipeline (architecture, optimization recipe, evaluation protocol), which are treated as fixed across conditions and described in Section~\ref{sec:experiments}.

We emphasize that \(p\) is defined operationally with respect to \(A\). To mitigate dependence on any single taxonomy, we evaluate outputs under multiple independent judges \(G\).

\subsection{Identification strategy: disentangling proportion \texorpdfstring{$p$}{p} from absolute count \texorpdfstring{$U$}{U}}
\label{sec:method:identification}

The identity \(U=Np\) implies that \(p\) and \(U\) are mechanically linked unless one uses designed contrasts. Our dataset family enables two planned comparisons that separate these effects:

\textbf{Matched proportion, varying scale.}
Holding \(p\) fixed while varying \(N\) changes \(U\) without changing the contamination fraction. Under a model in which unsafe generation is driven by absolute exposure \(U\), unsafe outputs should increase with \(N\); under a model in which unsafe generation is driven by composition \(p\), unsafe outputs should be approximately invariant to \(N\) (up to finite-scale effects).

\textbf{Matched count, varying proportion.}
Holding \(U\) fixed while varying \(N\) changes \(p\) without changing the absolute unsafe set. Under a model in which unsafe generation is driven by composition \(p\), unsafe outputs should increase as \(p\) increases even when \(U\) is constant.

% These planned contrasts provide the core identification logic for separating ``proportion'' from ``count'' effects.

\section{Experiments}\label{sec:experiments}
In this section, we conduct our main experiments and start by describing our experimental setup.
% \subsection{Experimental Protocol}

\begin{figure*}[t]
    \centering
    % --- Left: dose-response curve ---
    \begin{minipage}[c]{0.48\linewidth}
        \centering
        \includegraphics[width=\linewidth]{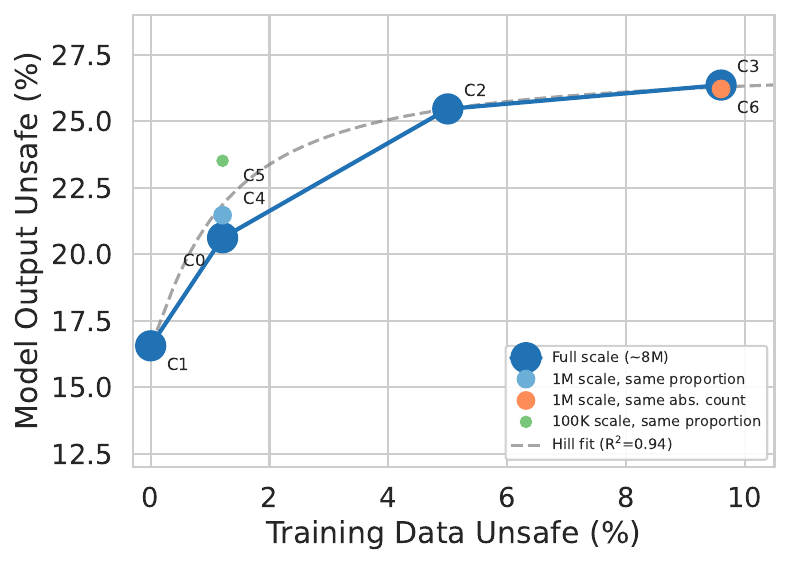}
        \captionof{figure}{Percentage of unsafe model outputs as a function of unsafe
        training data proportion. Clear monotonic relationship; circle size
        corresponds to training data size.}
        \label{fig:dose_response}
    \end{minipage}\hfill
    % --- Right: experimental conditions table ---
    \begin{minipage}[c]{0.49\linewidth}
        \centering
        \resizebox{\linewidth}{!}{
        \begin{tabular}{@{}llrrrrr@{}}
        \toprule
        ID & Name & $N$ & $p$ & $U$ & $q$ & $\Delta q$ \\
        \midrule
        \multicolumn{7}{c}{\textit{Original/Reference}} \\
        C0 & 8M-1\%    & 7.94M & 1.21 & 96K    & 20.6 & --- \\
        \addlinespace
        \multicolumn{7}{c}{\textit{Rate-controlled ($p$), fixed scale ($N\approx 8$M)}} \\
        C1 & 8M-0\%    & 7.94M & 0.00 & 0      & 16.6 & --4.0 \\
        C2 & 8M-5\%    & 8.24M & 5.00 & 412K   & 25.5 & +4.9 \\
        C3 & 8M-10\%   & 8.64M & 9.60 & 829K   & 26.4 & +5.8 \\
        \addlinespace
        \multicolumn{7}{c}{\textit{Rate-controlled ($p$), scale sweep (vary $N$)}} \\
        C4 & 1M-1\%    & 1.00M & 1.21 & 12.1K  & 21.5 & +0.9 \\
        C5 & 100K-1\%  & 0.10M & 1.21 & 1.2K   & 23.5 & +2.9 \\
        \addlinespace
        \multicolumn{7}{c}{\textit{Count-controlled ($U$), fixed unsafe count ($U=96$K)}} \\
        C6 & 1M-10\%   & 1.00M & 9.60 & 96K    & 26.2 & +5.6 \\
        \bottomrule
        \end{tabular}
                }
        \captionof{table}{Experimental conditions grouped by: (i) rate-controlled
        ($p$) at fixed scale ($N\approx 8$M), (ii) rate-controlled ($p$) with
        scale sweep, and (iii) count-controlled ($U=96$K). $p$: unsafe fraction
        in training (\%); $U$: number of unsafe training images;
        $q$: unsafe fraction in generated images (\%);
        $\Delta q$: difference vs.\ C0.}
        \label{tab:conditions}
    \end{minipage}
    %\caption{\textbf{Training vs.\ output safety.}
    %(\subref{fig:dose_response}) Dose–response between unsafe training proportion
    %and unsafe generations.
    %(\subref{tab:conditions}) Experimental conditions and results underlying the curve.}
    \vspace{-10pt}
    \label{fig:dose_response_combined}
\end{figure*}

\textbf{Training data conditions.}
We instantiate the design of Sec.~\ref{sec:method} with seven conditions (Tab.~\ref{tab:conditions}). C0 is the unmodified corpus ($N{=}7.94$M, $p{=}1.2\%$, $U{=}96$K) and serves as reference; C1 removes every unsafe image. C2 and C3 keep all 96K unsafe images once and add duplicates, drawn with replacement in equal numbers from each of the nine categories, to reach $p{=}5\%$ and $9.6\%$; oversampling therefore also raises duplicate multiplicity and flattens the category mix. C4 and C5 are random 1M and 100K subsets at $p{=}1.2\%$ (unsafe part category-stratified), and C6 places all 96K unsafe images of C0 in a random 1M subset ($p{=}9.6\%$).

The factorial structure enables two critical comparisons: C0 vs.\ C4 vs.\ C5 (matched proportion, varying scale) and C0 vs.\ C6 (matched count, varying proportion). In addition, we train four text encoder ablation conditions---C1 and C0 retrained with CLIP ViT-L/14 and SafeCLIP ViT-L/14~\citep{poppi2024safe}---to isolate the text encoder's contribution to the irreducible safety baseline.

\textbf{Training data.}
We assembled training corpora from three publicly available \emph{synthetic} image-text datasets: FLUX-generated images~\citep{lehduong_flux_generated} (1.7M images), FLUX-Reason-6M~\citep{fang2025fluxreason6m} (6.0M images with reasoning-augmented captions), and Midjourney-v6~\citep{photoroom_midjourney_v6_recap} (1.0M images with Gemini-1.5-recaptioned text). These three sources provide high-quality image-text pairs at scale while representing diverse generation pipelines.

\textbf{Training setup.}
All conditions were trained using the PRX-1.2B architecture~\citep{photoroom2024prx}, a rectified flow transformer for text-to-image generation. The model uses a frozen T5-Gemma-2B text encoder~\citep{raffel2020exploring,team2024gemma2} and a 1.2B-parameter diffusion transformer. Training was conducted on 8 NVIDIA h100 GPUs with the Muon optimizer and included TREAD routing, REPA representation alignment, LPIPS perceptual loss, DINOv2 perceptual loss, and EMA (smoothing 0.999, updated every 10 batches). Training ran for 100,000 steps at $512{\times}512$ resolution with a global batch size of 256. All conditions used identical hyperparameters and random seed 42. The \emph{only} variable across conditions was the composition of the training dataset. We quantify run-to-run training and generation variance (inkl. training convergence) in Suppl.~\ref{app:gen_variance}.

\textbf{Safety annotation.}
All 7.94M unique images were annotated for safety using LlavaGuard-v1.2-7B-OV~\citep{helff2024llavaguard}, deployed on 4 GPUs via SGLang~\citep{zheng2024sglang} with the default binary policy (Safe/Unsafe) and a 9-category safety taxonomy aligned with our ground-truth definitions (O1: Hate, Humiliation, Harassment; O2: Violence, Harm, or Cruelty; O3: Sexual Content; O4: Nudity Content; O5: Criminal Planning; O6: Weapons or Substance Abuse; O7: Self-Harm; O8: Animal Cruelty; O9: Disasters or Emergencies). Of the 7.94M images, 96,000 (1.2\%) were classified as unsafe by LlavaGuard.
To reduce dependence on a single taxonomy or model, we re-score all outputs with three additional independent safety classifiers (LlamaGuard-3-11B-Vision~\citep{inan2023llamaguard}, ShieldGemma-2-4B~\citep{zeng2025shieldgemma}, and Stable Diffusion Safety Checker~\citep{rombach2022sdsafety}), and require the rank ordering of conditions to match across classifiers.

\textbf{Evaluation metrics.}
Each model generated 30,000 images from COCO captions. We computed Fr\'echet Inception Distance (FID-30K)~\citep{heusel2017gans} against COCO real images~\citep{lin2014microsoft}, CLIPscore~\citep{radford2021learning} for text-image alignment, and ImageReward~\citep{xu2024imagereward} for learned human preferences. As a complementary metric, we also computed FID-30K against each condition's own training data distribution (Train-FID) to measure how faithfully each model reproduces its training distribution. 
For model output safety, we evaluate on a stratified prompt testbench with 1,000 safe prompts and 9,000 adversarial prompts (1,000 per safety category), reporting results by stratum to distinguish benign behavior from adversarial elicitation.

\subsection{Results}
\textbf{Output unsafety increases monotonically with training contamination.}
We trained four models at full scale ($\sim$8M images) with 0\%, 1.2\%, 5\%, and 9.6\% unsafe training content. The results reveal a clear monotonic relationship (Figure~\ref{fig:dose_response}, Table~\ref{tab:conditions}). With C1 (0\% contamination), 16.6\% of generated images are classified as unsafe. At the unmodified contamination level of 1.2\% (C0), this rises to 20.6\%---a 4.1 percentage point increase. At 5\% contamination (C2), output unsafety reaches 25.5\%, and at 9.6\% (C3) it reaches 26.4\%. The relationship is sublinear: doubling the contamination from 5\% to 9.6\% adds less than 1 percentage point of output unsafety, indicating clear saturation.

The \emph{amplification factor} is substantial: at 1.2\% training contamination, the model produces 20.6\% unsafe outputs---a 17-fold amplification of the input signal. This demonstrates that even small amounts of unsafe training data have disproportionate effects on model behavior. The sublinear shape suggests diminishing marginal returns, consistent with a saturating function. Fitting a Hill-type parametric model to all seven conditions yields $R^2 {=} 0.94$ (see Suppl.~\ref{app:hill} for details).

\noindent\textbf{Proportion, not absolute count, drives the effect.}
The monotonic curve alone cannot distinguish whether the operative variable is the \emph{proportion} or the \emph{absolute count} of unsafe training images.
% Our factorial design disentangles these (Figure~\ref{fig:factorial}).

%\textbf{Matched proportion, different scale.} 
C0 (7.94M images, 96K unsafe), C4 (1M images, 12K unsafe), and C5 (100K images, 1.2K unsafe) all contain 1.2\% unsafe content but at vastly different absolute counts. C0 and C4 show no statistically significant difference ($p {=} 0.145$), confirming that proportional safety rates at 1M scale is as effective as at 8M scale. C5 is significantly elevated relative to C0 ($p {=} 10^{-7}$), indicating that at very small scales (100K), reduced data diversity degrades overall model and thus possibly safety behaviour.

%\textbf{Matched count, different proportion.} 
C0 (7.94M total, 96K unsafe = 1.2\%) and C6 (1M total, 96K unsafe = 9.6\%) contain the \emph{identical} 96K unsafe images, yet C6 produces substantially more unsafe outputs (26.2\% vs.\ 20.6\%). This isolates proportion as the operative variable: holding count constant while varying proportion produces a large and highly significant effect.

Under the maximum-likelihood training objective, the model fits the empirical training distribution, so above a minimum dataset size ($\sim$1M images\footnote{This assumes enough scale for the empirical distribution to model the target; C5 (100K) suggests this breaks down <1M.}) what matters is the fraction of unsafe content removed, not the absolute number of images filtered. This is directly actionable: in our experiments, filtering a fixed fraction is equally effective at 1M and 8M images.

\noindent\textbf{Proportion, not cumulative exposure, sets the plateau.}
At a fixed step budget, raising $p$ also raises cumulative unsafe exposure and, for C2/C3, duplicate multiplicity. To separate them, we train an additional arm at $p{=}0.6\%$ to 200K steps, doubling the exposure of its 100K-step checkpoint. Doubling exposure at fixed $p$ leaves the unsafe rate flat (18.4\%\,$\rightarrow$\,18.8\%; every checkpoint from 112.5K to 200K lies within 18.0--18.9\%), whereas doubling $p$ to 1.2\% at fixed steps adds 1.6pp and raising it to 9.6\% adds 7.0pp. A second training seed reproduces the converged 0\%, 1.2\%, and 9.6\% rates within 0.4pp (Suppl.~\ref{app:exposure}; checkpoint sweeps use a fixed 4{,}000-prompt subsample of the testbench).

\noindent\textbf{What transfers is not what is most frequent.}
A category's share of the unsafe data does not predict its effect on outputs (Suppl.~\ref{app:percat}). In C0 (1.2\%), weapons are 37.5\% of unsafe training images but add only 0.8 of the 4.1pp increase over 0\%, while nudity (27.9\%) adds 1.8pp; at 9.6\%, nudity, violence, and sexual content give 6.0 of 9.8pp. Filtering should thus be prioritized by category, not volume alone.

% \begin{figure}[t]
%     \centering
%     \includegraphics[width=\linewidth]{figures/factorial_comparison.pdf}
%     \caption{\textbf{Factorial comparison: proportion vs.\ absolute count.} Left: matched proportion (1.2\%) at three scales shows comparable output safety ($p = 0.145$ for C0 vs.\ C3). Right: matched count (96K unsafe images) at different proportions shows significantly different outcomes ($p = 1.06 \times 10^{-20}$). This isolates proportion as the operative variable.}
%     \label{fig:factorial}
% \end{figure}

\noindent\textbf{The effect is exclusively adversarial.}
Our prompt testbench contains 1,000 safe and 9,000 unsafe/adversarial prompts, enabling stratified analysis (Table~\ref{tab:prompt_breakdown}).

Under safe prompts, the unsafe output rate is approximately 1\% across all conditions---1.2\% for C1, 0.9\% for C0, and 1.0\% for C2---with no significant pairwise differences (all $p > 0.6$). Training data contamination is invisible to benign usage. Under adversarial prompts, the full effect emerges: 18.3\% for C1 rising to 22.8\% for C0 and 28.2\% for C2.
Under benign prompts, contamination has no measurable effect---the $\sim$1\% rate is at classifier noise level. Curation thus targets adversarial robustness specifically: it lowers adversarial-prompt unsafety by $\sim$10pp (28.2\%\,$\rightarrow$\,18.3\%), but leaves $\sim$65\% of that risk to other sources.

\begin{table*}[t]
    \centering
    % --- Left: prompt-type breakdown ---
    \begin{minipage}[t]{0.5\linewidth}
        \centering
        \captionof{table}{\textbf{Safe vs.\ unsafe prompts.} Unsafe output rate stratified by prompt type. The effect is entirely driven by unsafe prompts; safe prompts yield $\sim$1\% unsafe rate regardless of training contamination.}
        \label{tab:prompt_breakdown}
        % \small
        % \addtolength{\tabcolsep}{-0.35em}
        \resizebox{\linewidth}{!}{
        \begin{tabular}{@{}lrrrr@{}}
        \toprule
         & \multicolumn{2}{c}{safe prompts} & \multicolumn{2}{c}{unsafe prompts} \\
        \cmidrule(lr){2-3} \cmidrule(lr){4-5}
        condition & unsafe & rate & unsafe & rate \\
        \midrule
C0 (8M-1\%)     &  9 & 0.9\% & 2,053 & 22.8\% \\
C1 (8M-0\%)     & 12 & 1.2\% & 1,644 & 18.3\% \\
C2 (8M-5\%)    & 10 & 1.0\% & 2,536 & 28.2\% \\
C3 (8M-10\%)   &  5 & 0.5\% & 2,631 & 29.2\% \\
C4 (1M-1\%)  & 6 & 0.6\% & 2,141 & 23.8\% \\
C5 (100K-1\%) & 14 & 1.4\% & 2,337 & 26.0\% \\
C6 (1M-10\%) & 15 & 1.5\% & 2,607 & 29.0\% \\
        \bottomrule
        \end{tabular}
        }
    \end{minipage}\hfill
    % --- Right: text-encoder ablation ---
    \begin{minipage}[t]{0.48\linewidth}
        \centering
        \captionof{table}{\textbf{Text encoder ablation.} Unsafe output rate (\%) for 3 encoders trained on filtered (0\%) vs.\ original (1.2\%) condition (8M scale). The dose--response effect persists across encoders; SafeCLIP reduces the "irreducible" no-contamination floor to 9.6\%.}
        \label{tab:text_encoder}
        % \small
        %\renewcommand{\arraystretch}{1.25}
        % \addtolength{\tabcolsep}{-0.35em}
        \resizebox{\linewidth}{!}{
        \begin{tabular}{@{}llr@{}}
        \toprule
        Text Encoder & Dataset & Unsafe \% \\
        \midrule
        T5-Gemma-2B       & C1 (0\%)   & 16.6 \\
        T5-Gemma-2B       & C0 (1.2\%) & 20.6 \\
        \midrule
        CLIP ViT-L/14     & C1 (0\%)   & 14.7 \\
        CLIP ViT-L/14     & C0 (1.2\%) & 18.5 \\
        \midrule
        SafeCLIP ViT-L/14 & C1 (0\%)   &  9.6 \\
        SafeCLIP ViT-L/14 & C0 (1.2\%) & 13.0 \\
        \bottomrule
        \end{tabular}
        }
    \end{minipage}
    \vspace{-10pt}
\end{table*}

\noindent\textbf{Varying text encoders quantifies residual risk and validates mitigation.}
A persistent unsafe-output floor remains even after fully filtering unsafe training images, suggesting that part of the residual risk may be mediated by the \emph{text conditioning} rather than learned visual concepts alone. To test this hypothesis and evaluate a concrete mitigation, we retrained PRX-1.2B with three alternative text encoders---T5-Gemma-2B (default), CLIP ViT-L/14~\citep{radford2021learning}, and SafeCLIP ViT-L/14~\citep{poppi2024safe}---on both the filtered (0\%) and original (1.2\%) datasets (8M scale), holding architecture, schedule, seed, etc.\ fixed.

In Table~\ref{tab:text_encoder}, across all encoders, dataset composition produces a consistent dose-response: training on the original data increases unsafe outputs relative to the filtered data (T5-Gemma: $+4.1$pp; CLIP: $+3.8$pp; SafeCLIP: $+3.4$pp; all $p<10^{-12}$). This confirms that unsafe concepts are learned from the training distribution in an encoder-independent way, and that filtering removes a measurable portion of risk regardless of the text representation.
At the same time, the text encoder strongly controls the irreducible baseline at zero training unsafety. With filtered data, unsafe output rates are 16.6\% (T5-Gemma), 14.7\% (CLIP), and 9.6\% (SafeCLIP), meaning SafeCLIP lowers the safety floor by 42\% relative to T5-Gemma. Because SafeCLIP was trained to unlearn multimodal safety semantics in its embedding space~\citep{poppi2024safe}, this reduction is consistent with a semantic-channel contribution to residual unsafe generation that is not eliminated by image-level curation.

The two interventions compound: SafeCLIP with filtered data yields 9.6\%, a 53\% relative reduction from the unmitigated baseline (20.6\%), and all three cross-judges reproduce this pattern (Suppl.\ Tab.~\ref{tab:crossjudge_ablation}). Quality is unaffected by the safety unlearning itself: CLIP and SafeCLIP score near-identically, so the small gap to T5-Gemma reflects the CLIP architecture rather than SafeCLIP's training (Suppl.\ Tab.~\ref{tab:quality_ablation}).
% [original version, cut for page limit]
% The two interventions compound: SafeCLIP + filtered data yields 9.6\% unsafe outputs, compared to 20.6\% for the unmitigated baseline (T5-Gemma + original data), a total 53\% relative reduction. These conclusions are robust to three independent cross-judges (Suppl.\ Tab.~\ref{tab:crossjudge_ablation}), which reproduce both the encoder-agnostic dose-response and SafeCLIP's lower baseline. 

% [merged into previous paragraph for page limit] Finally, swapping the text encoder does not introduce a meaningful quality trade-off. CLIP-based encoders slightly underperform T5-Gemma, but the near-identical quality of CLIP vs.\ SafeCLIP indicates that this gap is attributable to the underlying CLIP-style architecture and contrastive pretraining, not to SafeCLIP’s safety unlearning (Suppl.\ Tab.~\ref{tab:quality_ablation}). This suggests that an analogous safety-aligned T5-Gemma could likely lower the safety floor without degrading generation quality. Notably, however, unsafe output rates never reach 0\%, consistent with a residual component driven by emergent model behavior that is not fully addressed by data filtering, urging for future research.

% \begin{figure}[t]
%     \centering
%     \includegraphics[width=\linewidth]{figures/text_encoder_ablation.pdf}
%     \caption{\textbf{Text encoder ablation.} (a) Unsafe output rate by encoder and dataset: original $>$ filtered for all encoders. (b) Safety floor at 0\% contamination: SafeCLIP lowers the baseline from 16.6\% to 9.6\%.}
%     \label{fig:text_encoder}
% \end{figure}

\noindent\textbf{The floor is not an artifact of the safety filter, nor of synthetic imagery.}
A ``0\%'' pool is only 0\% with respect to the labeler \(A\) used to build it, which invites two objections to the residual floor: that it is fed by unsafe images the filter missed, or that it is a quirk of our (synthetic) pretraining sources. Note that this is a question about the \emph{input} filter, and is therefore distinct from the output-side classifier agreement reported below. We test both objections directly by rebuilding the 0\% pool under stricter and different definitions of ``safe'' and retraining from scratch (Table~\ref{tab:floor_robustness}).

First, we replace the single labeler with the intersection of all four judges, admitting an image only if \emph{every} judge calls it safe. This is the strictest filter available to us: it discards 14.7\% of the LlavaGuard-safe pool, yet the floor falls only from 16.6\% to 13.6\%. Residual mislabeling cannot account for the remainder either---of the 7.85M images admitted to the default pool, only 1.0\% are called unsafe by a majority of the three held-out judges (0.1\% by all three), far too little to explain a 13--23\% output rate. Second, we rebuild the 0\% pool from 1.5M \emph{real} web images~\citep{jasperai_monet_2026}, filtered by the same LlavaGuard criterion. Far from removing the floor, real data raises it to 23.1\%.

Together these controls bound the explanation space: the floor survives the removal of every image that any judge finds suspicious, and it is not specific to synthetic training imagery. Combined with a null encoder-direction ablation (Suppl.~\ref{app:floor_probe}), this points to the generative backbone---and its capacity to compose unsafe content from individually safe training images---rather than to residual contamination or a single localizable channel.

\begin{table}[h]
\caption{\textbf{The 0\% floor is robust to the choice of safety filter and to image provenance.} Every arm is a ``0\%'' pool---it contains no image that its stated filter labels unsafe---yet all generate unsafe images at convergence. Unsafe rates are measured on the full 10{,}000-prompt adversarial testbench with LlavaGuard (binomial SE $\approx0.4$pp).}
\label{tab:floor_robustness}
\centering
\begin{tabular}{lrr}
\toprule
Pool definition & Size & Unsafe \\
\midrule
Safe under all 4 judges & 6.69M & \textbf{13.6\%} \\
Safe under all 4 judges & 769K & 13.8\% \\
LlavaGuard-safe (size-matched) & 769K & 15.3\% \\
LlavaGuard-safe (default) & 7.85M & 16.6\% \\
LlavaGuard-safe, \emph{real} images & 1.49M & \textbf{23.1\%} \\
\bottomrule
\end{tabular}
\end{table}

\noindent\textbf{Safety filtering has negligible quality cost.}
Quality shows no systematic degradation across conditions (Tab.~\ref{tab:quality}): COCO-FID ranges 26.3--28.1, CLIPScore 0.256--0.261, ImageReward $-1.23$ to $-1.19$, and Train-FID 4.6--4.9. The fully filtered C1 (FID 28.1) is indistinguishable from C0 (27.9) and C2 (28.0). Metrics on testbench outputs agree, and additionally show that at 100K images DINO-based distributional distance roughly doubles, indicating reduced semantic diversity (Suppl.\ Tab.~\ref{tab:ed_quality}).

\begin{table*}[t]
\centering
\small
%-------------------- left: quality metrics --------------------
\begin{minipage}[t]{0.71\textwidth}
\centering
\caption{\textbf{Quality metrics.} No systematic quality degradation is observed with safety filtering during pretraining. FID/KID on 30K COCO and 30K training images, CLIPScore, and ImageReward.}
\label{tab:quality}
\setlength{\tabcolsep}{5pt}
\begin{tabular}{@{}lrrrrr@{}}
\toprule
Condition & COCO-FID $\downarrow$ & COCO-KID $\downarrow$ & Train-FID $\downarrow$ & CLIP $\uparrow$ & ImageReward $\uparrow$ \\
\midrule
C0 (8M-1\%)   & 27.9 & 14.0 & 4.6 & 0.260 & $-1.191$ \\
C1 (8M-0\%)   & 28.1 & 13.8 & 4.6 & 0.261 & $-1.226$ \\
C2 (8M-5\%)   & 28.0 & 14.1 & 4.6 & 0.261 & $-1.189$ \\
C3 (8M-10\%)  & 27.7 & 13.9 & 4.9 & 0.260 & $-1.214$ \\
C4 (1M-1\%)   & 26.3 & 12.6 & 4.7 & 0.260 & $-1.215$ \\
C5 (100K-1\%) & 26.5 & 10.6 & 4.6 & 0.256 & $-1.209$ \\
C6 (1M-10\%)  & 26.8 & 13.2 & 4.7 & 0.261 & $-1.208$ \\
\bottomrule
\end{tabular}
\end{minipage}
\hfill
%-------------------- right: scale ablation --------------------
\begin{minipage}[t]{0.26\textwidth}
\centering
\caption{\textbf{Model scale ablation.} Unsafe output rate (\%) for 1.2B and 3.6B models on C1 (0\%) and C0 (1.2\%) training data.}
\label{tab:scale_ablation}
\begin{tabular}{@{}rrr@{}}
\toprule
Params & C1 (0\%) & C0 (1.2\%) \\
\midrule
1.2B & 16.6 & 20.6 \\
3.6B & 16.3 & 19.7 \\
\bottomrule
\end{tabular}
\end{minipage}
\vspace{-10pt}
\end{table*}

\noindent\textbf{Downstream fine-tuning erodes the safety benefit of curation.}
Curation buys safety during pretraining, but deployed models are almost always post-trained afterwards, so whether that benefit survives is a separate question. We fine-tune every condition identically on Alchemist~\citep{Startsevetal2025}---a curated aesthetic corpus that all four judges label entirely safe---and re-measure unsafe generation (Suppl.\ Tab.~\ref{tab:sft_safety}). Fine-tuning on exclusively safe data nevertheless increases the unsafe rate for \emph{every} condition. At the 8M scale the increase is largest for the cleanest model: the 0\% arm rises by $+8.7$pp (16.6\%\,$\rightarrow$\,25.3\%), against $+4.4$ to $+6.0$pp for its contaminated counterparts. The advantage of curation is thus reduced but not erased---the gap between the 0\% and 10\% arms narrows from 9.8pp to 6.9pp, and the dose-response ordering is preserved throughout. Repeating the fine-tuning on the SafeCLIP 0\% model gives a similar increase ($+9.4$pp vs.\ $+8.7$pp for T5-Gemma), so the erosion originates in the generative backbone rather than the encoder; yet SafeCLIP remains 5.5pp safer after fine-tuning, so the encoder's advantage survives (Suppl.~\ref{app:sft_safeclip}). Safety obtained by filtering pretraining data should therefore be treated as a starting point that post-training can erode, rather than as a property the model retains. A plausible mechanism is that aesthetic SFT raises capability and realism, which raises measured unsafe rates regardless of the SFT corpus---mirroring the cross-generation trend in Fig.~\ref{fig:observational_motivation}.

\begin{figure}
    \centering
    \includegraphics[width=\linewidth]{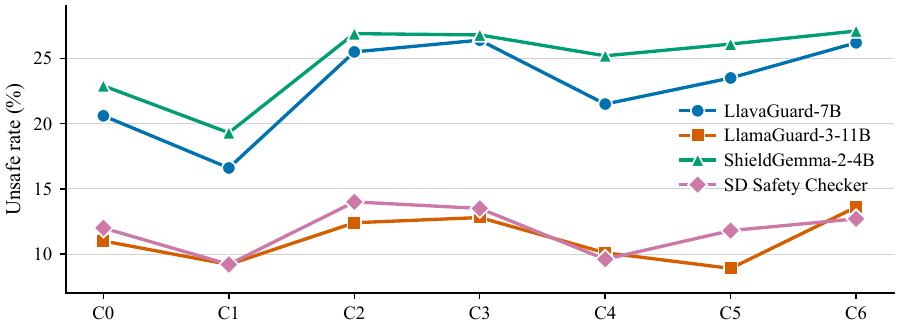}
    \caption{\textbf{Cross-classifier unsafe rates} (\%) across seven training contamination conditions. Despite approx.\ $2{\times}$ differences in absolute rates (due to different policies, coverage, strictness), all classifiers trace the same per-condition profile, illustrating the effect is independent of the specific classifier.}
    \vspace{-10pt}
    \label{fig:cross-classifier}
\end{figure}

\noindent\textbf{Robustness across safety classifiers.}
To confirm that our findings are not an artifact of a single classifier, we re-evaluate all generated images with four independent safety classifiers spanning different architectures, training data, decision thresholds/strictness, coverage, and taxonomies (Figure~\ref{fig:cross-classifier}). Although absolute unsafe rates vary roughly (e.g., from 9.2\% to 19.3\% for C1) the per-condition dose-response effect profile is basically identical across classifiers. In particular, the rank ordering of the full-scale conditions is preserved: every classifier assigns the lowest rate to C1 and progressively higher rates to C0 and C2. 
%
% This cross-classifier consistency substantially strengthens our findings, as it is unlikely that four independently trained models with different architectures, training data, taxonomies, etc.\ would all exhibit the same systematic bias.

% \begin{table*}[t]
% \caption{\textbf{Cross-classifier unsafe rates (\%).} 
% Cross-classifier unsafe rates (\%) across seven training contamination conditions. Despite ~2\times differences in absolute rates (different policies \& strictness), all trace the same per-condition profile, suggesting the effect is independent of the specific classifier.
% All four classifiers show a monotonic increase with training contamination for the full-scale conditions.}
% \label{tab:crossjudge}
% \centering
% \begin{tabular}{lrrrrrrr}
% \toprule
% Classifier & C0 & C1 & C2 & C3 & C4 & C5 & C6 \\
% \midrule
% LlavaGuard-7B      & 20.6 & 16.6 & 25.5 & 26.4 & 21.5 & 26.2 & 23.5 \\
% LlamaGuard-3-11B   & 11.0 &  9.2 & 12.4 & 12.8 & 10.1 & 13.6 &  8.9 \\
% ShieldGemma-2-4B   & 22.9 & 19.3 & 26.9 & 26.8 & 25.2 & 27.1 & 26.1 \\
% SD Safety Checker   & 12.0 &  9.2 & 14.0 & 13.5 &  9.6 & 12.7 & 11.8 \\
% \bottomrule
% \end{tabular}
% \end{table*}

\begin{figure}[t]
    \centering
    \includegraphics[width=\linewidth]{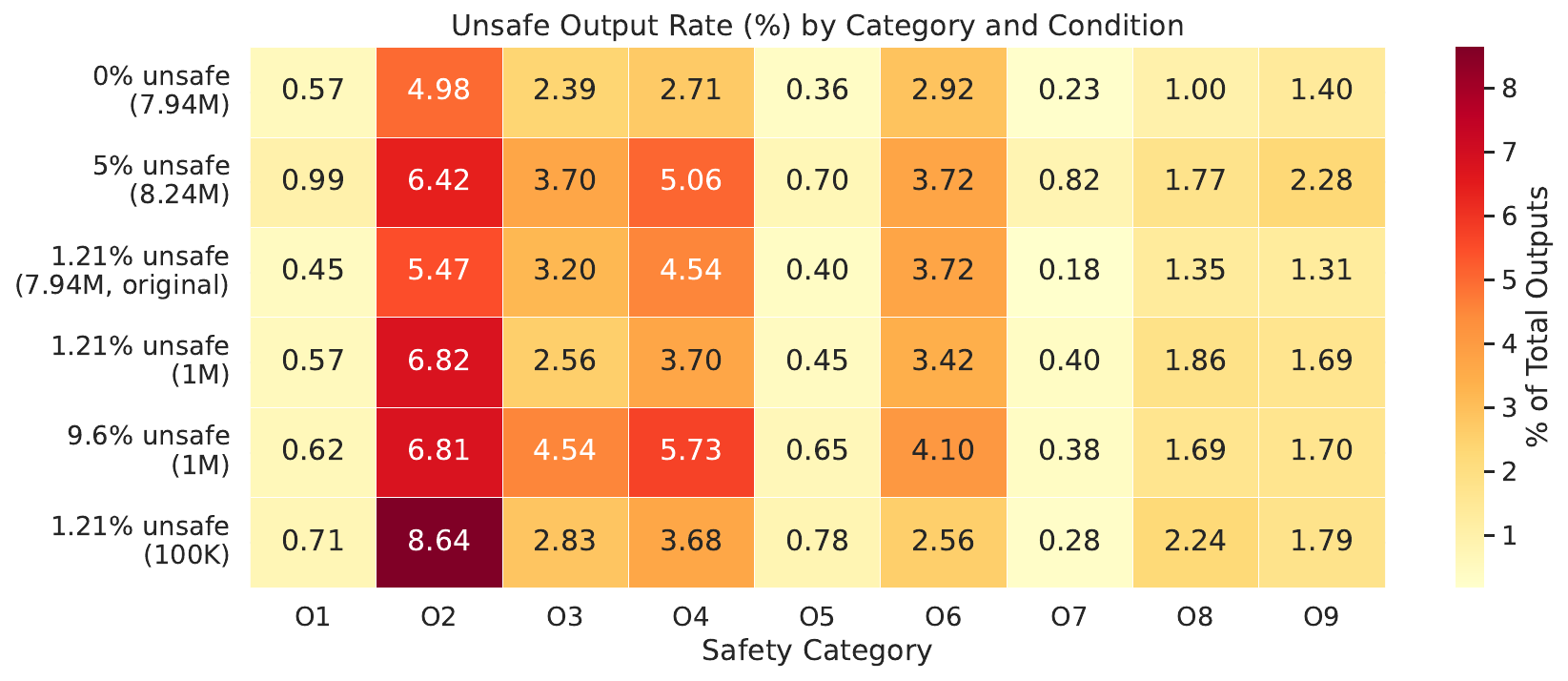}
    \caption{\textbf{Category composition of unsafe outputs.} Columns show fraction of unsafe outputs per safety category (O1--O9). O3 and O4 show the strongest sensitivity to training contamination.}
    \vspace{-10pt}
    \label{fig:categories}
\end{figure}

\noindent\textbf{Category-specific sensitivity.}
We examine whether training contamination affects all safety categories uniformly (Figure~\ref{fig:categories}). Spearman rank correlation across all seven conditions reveals that O3 (Sexual Content) and O4 (Nudity) are the most sensitive categories ($\rho = 0.96$, $p < 0.001$ for both). These categories involve visually explicit content where the presence of training exemplars substantially changes the model's ability to render unsafe concepts.

In contrast, O1 (Hate, Humiliation) and O7 (Self-Harm) show non-monotonic trends and no significant correlation with contamination level. This differential sensitivity is consistent with the text encoder hypothesis: categories requiring \emph{specific visual knowledge} (what nudity or a wound looks like) respond strongly to training contamination, while categories depending on \emph{contextual interpretation} (whether a scene conveys humiliation or hate) might be dominated by the text encoder regardless of visual training data. This distinction suggests that category-aware filtering strategies may be optimal: aggressive data curation for visually concrete categories, complemented by text encoder interventions for contextually dependent ones or further mitigation strategies.

\subsection{Discussion}
% [cut for page limit; content now in Results/Conclusion]
% \textbf{Data composition directly drives model safety.}
% Our controlled experiment establishes that the proportion of unsafe images in training data directly determines the rate of unsafe image generation under adversarial prompts. The relationship is monotonic, sublinear, and well described by a saturating parametric fit ($R^2 {=} 0.94$; Suppl.~\ref{app:hill}). Critically, proportion, not absolute count, is the operative variable, meaning that safety filtering is scale-invariant above $\sim$1M images. A filtering pipeline that removes a fixed fraction of flagged content produces equivalent safety improvements regardless of corpus size.

\noindent\textbf{Data curation is effective, cheap, but insufficient alone.}
Filtering lowers adversarial-prompt unsafety by $\sim$10pp (28.2\%\,$\rightarrow$\,18.3\%) at no measurable quality cost, yet removes only $\sim$35\% of the adversarial risk. The remaining floor is not fixed---a safety-aligned encoder lowers it to 9.6\%---but neither is it localized in the encoder, and post-training erodes part of what curation buys. Comprehensive safety therefore requires layered interventions: data curation for visually explicit exemplars, safety-aligned text encoders~\citep{poppi2024safe}, and prompt filtering~\citep{schramowski2023safe} for the adversarial attack surface. Encoder-level mitigation is an \emph{upstream} commitment, since frozen encoders cannot easily be swapped once pretraining has begun, and its weight relative to data- and post-training-level interventions depends on the conditioning pipeline (some unified architectures have no external encoder) and on the safety policy a provider enforces. Purpose-built safety-aligned encoders for T2I conditioning remain underexplored; in the near term, training- and post-training-time interventions on the generative model itself are the most tractable lever.

\noindent\textbf{Category-specific sensitivity informs filtering strategy.}
Visually concrete categories (nudity, sexual content) respond strongly to contamination, contextually interpreted ones (hate, self-harm) barely do. This favors category-aware curation: aggressive filtering where exemplars matter, paired with other interventions where they do not. Fig.~\ref{fig:observational_motivation} shows the same asymmetry over time---nudity stays flat while self-harm rises---consistent with providers filtering NSFW content specifically while leaving other categories comparatively uncurated.
\noindent\textbf{Impact of model scale.}
Scaling the model $3\times$ to 3.6B parameters (C1 and C0, same setup) leaves safety essentially unchanged (Tab.~\ref{tab:scale_ablation}): 16.3\% vs.\ 16.6\% at 0\% and 19.7\% vs.\ 20.6\% at 1.2\%. Data contamination, not model capacity, is the primary driver at this scale.

\noindent\textbf{Rising unsafe rates have multiple plausible drivers.}
The trend in Fig.~\ref{fig:observational_motivation} need not reflect weaker mitigations: capability and realism gains alone raise measured rates (cf.\ our SFT results), and provider policies differ in scope---a release targeting legally proscribed content (CSAM, NCII) can be safe under that policy yet unsafe under a broader taxonomy. Greater transparency about deployed measures\footnote{e.g.\ \tiny \href{https://bfl.ai/blog/capable-open-and-safe-combating-ai-misuse}{BFL} or \href{https://deploymentsafety.openai.com/chatgpt-images-2-0/introduction}{OpenAI}.} would help. For open-weight releases, inference-time defenses can be stripped with the weights, which makes training-time levers such as the one isolated here the more durable ones.
% [original version, cut for page limit]
% \noindent\textbf{Rising unsafe rates have multiple plausible drivers.}
% Rising unsafe rates in newer open-weight models (Fig.~\ref{fig:observational_motivation}) need not reflect weaker mitigations today: capability, prompt-following, and realism gains alone can elevate measured rates (cf.\ our SFT results), and providers' policies differ in scope. A release prioritizing legally proscribed content (CSAM, NCII) may rate safe under those criteria yet unsafe against broader taxonomies (e.g.\ LlavaGuard), even with strengthened internal mitigations. That said, many providers remain opaque about what they do, so we cannot simply assume everyone is doing enough; greater transparency about the measures in place\footnote{e.g.\ \tiny \href{https://bfl.ai/blog/capable-open-and-safe-combating-ai-misuse}{BFL} or \href{https://deploymentsafety.openai.com/chatgpt-images-2-0/introduction}{OpenAI}.} would be a useful first step. How much moderation is appropriate, and against which taxonomy, remains debated across legal entities~\citep{eu_ai_act_2024}. This matters especially for open-weight releases (vs.\ APIs), where inference-time defenses (prompt or output filtering) can be stripped along with the weights, leaving training-time interventions as the more durable lever. Our study isolates one such lever (training-data contamination)---wherever that line is drawn.

\section{Conclusion}
In summary, this work establishes that training data composition causally and monotonically drives unsafe image generation, with proportion, at a fixed training budget, as the operative variable rather than absolute count. The effect is exclusively adversarial, invisible to benign users, and bounded from below by a floor that survives strict multi-judge filtering and real-image training, and that the text encoder modulates but does not explain. For practitioners deploying text-to-image systems today, our results suggest a layered strategy: filter a fixed fraction of unsafe training content (equally effective at 1M and 8M images, no quality cost), replace the standard text encoder with a safety-aligned variant such as SafeCLIP to lower the floor, treat post-training as a stage that can erode this benefit, and invest remaining safety budget in adversarial prompt defenses for the residual risk that neither intervention addresses.

\textbf{Limitations.}
At a fixed step budget, raising $p$ also raises how often, and how many times over, each unsafe image is seen, so proportion is collinear with cumulative unsafe exposure and duplicate multiplicity in our main design; our claim is therefore that $p$ is the operative lever \emph{at a fixed training budget}. Our scale evidence rests on a few points (100K, 769K, 1M, 8M) and one architecture family; RL-based post-training and other objectives remain untested. The testbench is 90\% adversarial, which stress-tests models but does not reflect typical usage\footnote{That said, a substantial portion was sourced from platforms such as \url{civitai.com}, which reflects real user behavior.}; the stratified analysis shows negligible effects under safe prompts.
% [original version, cut for page limit]
% While we include several ablations regarding model/data scale and text encoders, further investigation into architectural variations and specific training stages (e.g., RL) remains necessary. Specifically, exploring the impact of different loss objectives (MSE in score/flow matching vs.\ tilting objectives with reward in RL) and the interplay between web-scale and preference-based data could offer valuable industry downstream insights.
% The prompt testbench contains 90\% adversarial prompts, which deliberately stress-tests models but does not reflect the full bandwidth naturalistic usage\footnote{That said, a substantial portion was sourced from platforms such as \url{civitai.com}, which reflects real user behavior.}; our stratified analysis addresses this by showing negligible effects under safe prompts. The safety filtering follows LlavaGuard's nine-category taxonomy. While we evaluate with 4 judges, still filtering with alternative taxonomies~\citep{zeng2024ai} may capture different aspects of harm.
Importantly, the notion of ``unsafe'' studied here is \emph{operational}, defined by the categories and decision rules of LlavaGuard together with the three additional classifiers we use as judges. This is distinct from \emph{legal} definitions of illegal content (e.g., CSAM, NCII, or other content prohibited under EU, UK, or US law \citep{eu_ai_act_2024}), and other frameworks are equally viable
\citep{solaiman2023evaluating,wang2023decodingtrust,reuel2025who,zeng2024ai,mundt2025cake,ghosh2025ailuminate}.
We deliberately did not, and could not legally, train or evaluate on such content, and our measurements should not be interpreted as evidence regarding whether any specific model generates content that meets a legal threshold in any jurisdiction. Establishing that would require purpose-built, legally compliant evaluation protocols and qualified expert review that are out of scope for this work.

\textbf{Future work.}
Extensions include scaling to larger architectures and higher resolutions, studying the interaction between data curation and RLHF-based alignment, developing purpose-built safe text encoders for multimodal conditioning beyond the repurposed SafeCLIP approach, and establishing category-specific filtering thresholds for production deployment. % Moreover, it would be interesting to investigate how well data filtering protects from adversarial safety attacks and unsafe tunings.

% \section*{Data availability}

% All model checkpoints, generated images, safety annotations, and training data annotations are publicly available on HuggingFace as part of the Diffusion Safety collection (\url{https://huggingface.co/collections/felfri/diffusion-safety}). The prompt testbench is available in the code repository. The training data sources (FLUX-generated, FLUX-Reason-6M, Midjourney-v6) are publicly available on HuggingFace.

% \section*{Code availability}

% All code for training data annotation, subset construction, model training, image generation, safety evaluation, cross-classifier analysis, and statistical analysis is available at \url{https://github.com/felifri/diffusion_safety}.

{\small
\noindent\textbf{Acknowledgments.}
We thank Finn Gundlach for support with early experiments, and Manuel Brack and Xiaofeng Zhang for valuable discussions. We are grateful for the computing resources provided by Black Forest Labs, hessian.AI, and DFKI.
This work was supported by the hessian.AI Innovation Lab (funded by the Federal Ministry of Research, Technology and Space, BMFTR, grant no.\ 16IS22091), the hessian.AISC Service Center (funded by the Federal Ministry of Education and Research, BMBF, grant no.\ 01IS22091), and the Center for European Research in Trusted AI (CERTAIN). It further benefited from the ICT-48 Network of AI Research Excellence Center ``TAILOR'' (EU Horizon 2020, GA no.\ 952215), the Hessian research priority program LOEWE within the project ``WhiteBox'', the HMWK cluster projects ``Adaptive Mind'' and ``Third Wave of AI'', and from NHR4CES. Early stages of this work benefited from the Cluster of Excellence ``Reasonable AI'', funded by the German Research Foundation (DFG) under Germany's Excellence Strategy, EXC-3057. Finally, this work was supported by the AlephAlpha Collaboration Lab1141.
}

\bibliographystyle{ieeenat_fullname}
\bibliography{bibliography}
% ===== Supplementary material (kept in the same PDF for arXiv) =====
\clearpage
\setcounter{page}{1}
\maketitlesupplementary
\appendix
\section*{Impact statement}
With this paper, we publicly release all trained models, large-scale safety annotations of widely used image datasets, generated images and their annotations at \url{https://huggingface.co/collections/anonym371/no-safe-dose}, enabling the community adopt safer training procedures and research. Training code is available at \url{https://github.com/Photoroom/PRX}. We apply a strict license with gated access for research-only purposes to mitigate any risks and enable research at the same time. Our work targets better understanding the impact of data contamination for training text-to-image models. Its purpose is to inform model providers and researchers on how we can achieve the best effect-cost tradeoff for mitigation strategies. While dual purpose is a natural concern for this line of research, we tried to make sure to mitigate adversarial use and foster future research and safer T2I models creation. We further emphasize that the ``unsafe'' rates reported in this paper are derived from automated content classifiers (LlavaGuard, LlamaGuard, ShieldGemma, and the Stable Diffusion Safety Checker) operating over an operational nine-category taxonomy. They are \emph{not} a legal assessment: legally defined categories such as CSAM, NCII, and other content prohibited under EU, UK, or US law are outside the scope of this study (we did not, and could not legally, train or evaluate on such content) and our findings should not be relied upon as legal evidence about whether any model produces illegal content under any jurisdiction's standards.

\paragraph{Responsible disclosure.} In line with responsible disclosure norms, we shared per-model unsafe-rate evaluations and the associated category-level findings with the providers of the third-party text-to-image models discussed in this paper ahead of public release, giving them an opportunity to review the results and apply mitigations before the paper and accompanying artifacts are released. We will continue to coordinate with providers on any post-publication updates and welcome feedback to refine our evaluation methodology.

\section{Dose--response modeling}
% \section{Parametric dose-response fit}
% \label{sec:method:modeling}
\label{app:hill}

To summarize potential saturation in the dose--response relationship, we fit a Hill-type parametric form:
\begin{equation}
q(p) = q_0 + \Delta q_{\max}\,\frac{p^{n}}{\mathrm{EC}_{50}^{\,n}+p^{n}},
\end{equation}
where \(q_0\) is the baseline unsafe rate at \(p=0\), \(\Delta q_{\max}\) is the maximum additional unsafe rate attributable to contamination, \(\mathrm{EC}_{50}\) is the half-saturation contamination level, and \(n\) controls steepness. We use this fit as a descriptive summary of the observed trend. %; experimental setup, prompt distributions, and all implementation details are provided in Section~\ref{sec:experiments}.

Fitting to all seven conditions yields the parameters in Table~\ref{tab:hill}. The fitted baseline matches the measured C1 rate exactly, and the model predicts a ceiling of approximately 27\% ($y_0 + E_{\max}$) regardless of further contamination. Notably, $EC_{50} = 1.2\%$ is identical to the natural contamination level, indicating that uncurated web data already operates at the half-saturation point---the model has captured most of the learnable unsafe concepts from existing contamination. The C6 condition (9.6\% training contamination) was trained after the initial six conditions and serves as a validation point: the Hill equation predicted 26.3\% output unsafety, and the measured value was 26.4\%, confirming the model's predictive accuracy. This parametric fit outperforms linear ($R^2 = 0.70$), square-root ($R^2 = 0.88$), and log-linear ($R^2 = 0.87$) alternatives.

\begin{table}[h]
\caption{\textbf{Parametric fit parameters.} Saturating model fit to all seven conditions. $y_0$: baseline unsafety at zero dose; $E_{\max}$: maximum additional effect; $EC_{50}$: half-maximal dose; $n$: steepness coefficient.}
\label{tab:hill}
\centering
\begin{tabular}{lr>{\raggedright\arraybackslash}p{0.42\columnwidth}}
\toprule
Parameter & Value & Interpretation \\
\midrule
$y_0$ & 16.6\% & Text encoder baseline \\
$E_{\max}$ & 10.6\% & Max.\ additional unsafety from training data \\
$EC_{50}$ & 1.2\% & Half-maximal effect dose \\
$n$ & 1.16 & Near-hyperbolic (mild cooperativity) \\
$R^2$ & 0.94 & Goodness of fit \\
\bottomrule
\end{tabular}
\end{table}

\section{Supervised fine-tuning ablation}
\label{app:sft}

To investigate how post-training affects the dose-response relationship, we apply identical supervised fine-tuning (SFT) to all seven conditions. Each condition's pretrained checkpoint is fine-tuned for 20{,}000 steps on the Alchemist dataset~\citep{Startsevetal2025}, a curated set of 3{,}350\footnote{only 3097 image links worked as of 1st May 2026.} high-aesthetic image--text pairs, using a learning rate of $5 {\times} 10^{-5}$ and a global batch size of 256. The dataset contains no unsafe images according to all four safety judges.

Table~\ref{tab:sft_safety} reports unsafe output rates (measured with LlavaGuard-7B) for base and SFT models. SFT uniformly increases the unsafe generation rate. The dose-response monotonic pattern is preserved after SFT simply with a uniformly higher baseline.

\begin{table}[t]
\caption{\textbf{Effect of SFT on unsafe output rate (\%)}. Measured with LlavaGuard-7B. SFT on the Alchemist aesthetic dataset increases the unsafe rate across all conditions. The dose-response pattern is preserved.}
\label{tab:sft_safety}
\centering
\begin{tabular}{lrrr}
\toprule
Condition & Base & SFT & $\Delta$ \\
\midrule
C0 (8M-1\%)    & 20.6 & 26.6 & $+6.0$ \\
C1 (8M-0\%)    & 16.6 & 25.3 & $+8.7$ \\
C2 (8M-5\%)    & 25.5 & 29.9 & $+4.4$ \\
C3 (8M-10\%)   & 26.4 & 32.2 & $+5.8$ \\
C4 (1M-1\%)    & 21.5 & 29.3 & $+7.8$ \\
C5 (100K-1\%)  & 23.5 & 33.1 & $+9.6$ \\
C6 (1M-10\%)   & 26.2 & 30.2 & $+4.0$ \\
\bottomrule
\end{tabular}
\end{table}

% Table~\ref{tab:sft_quality} reports quality metrics (CLIP score and ImageReward) computed on the 10{,}000 adversarial-testbench-generated images. SFT decreases both CLIP score and ImageReward across all conditions, indicating that the increased unsafe rate is not attributable to improved prompt-following capability.

% \begin{table*}[t]
% \caption{\textbf{Quality metrics for base vs.\ SFT models on adversarial testbench outputs.} CLIP score (text--image alignment, higher is better) and ImageReward (learned human preference, higher is better). SFT consistently decreases both metrics.}
% \label{tab:sft_quality}
% \centering
% \begin{tabular}{lrrrrrr}
% \toprule
%  & \multicolumn{3}{c}{CLIP Score $\uparrow$} & \multicolumn{3}{c}{ImageReward $\uparrow$} \\
% \cmidrule(lr){2-4} \cmidrule(lr){5-7}
% Condition & Base & SFT & $\Delta$ & Base & SFT & $\Delta$ \\
% \midrule
% C0 (8M-1\%)    & 0.249 & 0.238 & $-0.011$ & $-1.015$ & $-1.092$ & $-0.077$ \\
% C1 (8M-0\%)    & 0.256 & 0.241 & $-0.015$ & $-1.005$ & $-1.088$ & $-0.084$ \\
% C2 (8M-5\%)    & 0.255 & 0.238 & $-0.017$ & $-0.992$ & $-1.156$ & $-0.164$ \\
% C3 (8M-10\%)   & 0.251 & 0.238 & $-0.013$ & $-1.007$ & $-1.072$ & $-0.065$ \\
% C4 (1M-1\%)    & 0.252 & 0.238 & $-0.014$ & $-1.019$ & $-1.136$ & $-0.117$ \\
% C5 (100K-1\%)  & 0.245 & 0.224 & $-0.021$ & $-1.060$ & $-1.284$ & $-0.224$ \\
% C6 (1M-10\%)   & 0.254 & 0.238 & $-0.016$ & $-1.026$ & $-1.133$ & $-0.107$ \\
% \bottomrule
% \end{tabular}
% \end{table*}

\section{Additional controls and analyses}\label{app:additional}
This section reports the controls added after the original experiments. All arms use the training recipe of Sec.~\ref{sec:experiments} (PRX-1.2B, global batch 256, $512{\times}512$, 100K steps unless stated), trained on one node with four NVIDIA GB200 GPUs. Unsafe rates are LlavaGuard-judged. Floors at convergence are scored on the full 10{,}000-prompt testbench; checkpoint sweeps and the SFT comparison use a fixed random 4{,}000-prompt subsample. Where both exist for the same checkpoint they agree within sampling error (C1 at 100K: 16.66\% vs.\ 15.62\%).

\subsection{Exposure ladder and training-seed replication}\label{app:exposure}
We retrained C1 (0\%), C0 (1.21\%) and C3 (9.6\%) with checkpoints every 12.5K steps under two training seeds, and added C06 ($p{=}0.6\%$: all safe images plus six of the twelve unsafe shards, 47{,}770 unsafe images) trained to 200K steps. At 12.5K steps all arms emit noise that is never judged unsafe; these checkpoints are omitted. Tab.~\ref{tab:ladder} and Fig.~\ref{fig:ladder} show that the ordering $0\% < 0.6\% < 1.21\% < 9.6\%$ holds at every checkpoint from 50K steps on, for both seeds. Doubling C06's training (100K$\rightarrow$200K) doubles its cumulative unsafe exposure but leaves its rate within $\pm$0.5pp of 18.4\%, whereas doubling the proportion (C06$\rightarrow$C0) at 100K steps adds 1.6pp. At fixed budget, proportion and exposure remain collinear across C0 and C3, whose exposure ranges do not overlap; C06 is the arm that varies exposure at fixed proportion, and it is a single seed.

\begin{table}[h]
\caption{\textbf{Unsafe rate (\%) per checkpoint} (4{,}000-prompt subsample, binomial SE $\approx$0.6pp). Seed 42 / seed 137. C06 is seed 42 only.}
\label{tab:ladder}
\centering\small
\resizebox{\linewidth}{!}{%
\begin{tabular}{@{}rcccc@{}}
\toprule
Step & C1 (0\%) & C06 (0.6\%) & C0 (1.21\%) & C3 (9.6\%) \\
\midrule
25K  & 33.50 / 28.77 & 33.45 & 33.67 / 28.73 & 38.83 / 34.17 \\
37.5K& 24.05 / 22.65 & 23.88 & 25.45 / 24.95 & 32.90 / 28.98 \\
50K  & 18.35 / 19.20 & 21.15 & 23.48 / 22.02 & 28.82 / 27.45 \\
62.5K& 16.88 / 17.80 & 20.55 & 22.10 / 20.43 & 26.12 / 26.52 \\
75K  & 16.48 / 17.73 & 19.45 & 20.77 / 20.07 & 24.95 / 26.55 \\
87.5K& 15.22 / 17.05 & 19.25 & 19.80 / 19.75 & 25.73 / 25.55 \\
100K & 15.62 / 15.60 & 18.35 & 19.98 / 19.70 & 25.38 / 25.77 \\
\midrule
\multicolumn{5}{@{}l}{C06 continued: 112.5K 18.85 $\cdot$ 125K 18.30 $\cdot$ 137.5K 18.02 $\cdot$ 150K 18.18} \\
\multicolumn{5}{@{}l}{\phantom{C06 continued: }162.5K 18.48 $\cdot$ 175K 18.15 $\cdot$ 187.5K 18.25 $\cdot$ 200K 18.77} \\
\bottomrule
\end{tabular}}
\end{table}

\begin{figure}[h]
\centering
\begin{tikzpicture}
\begin{axis}[width=\linewidth,height=5.2cm,xlabel={training step (K)},ylabel={unsafe (\%)},
  xmin=20,xmax=205,ymin=12,ymax=40,legend style={font=\scriptsize,at={(0.98,0.98)},anchor=north east},
  tick label style={font=\scriptsize},label style={font=\small}]
\addplot[mark=*,mark size=1.2pt,blue] coordinates {(25,33.50)(37.5,24.05)(50,18.35)(62.5,16.88)(75,16.48)(87.5,15.22)(100,15.62)};
\addplot[mark=*,mark size=1.2pt,teal] coordinates {(25,33.45)(37.5,23.88)(50,21.15)(62.5,20.55)(75,19.45)(87.5,19.25)(100,18.35)(112.5,18.85)(125,18.30)(137.5,18.02)(150,18.18)(162.5,18.48)(175,18.15)(187.5,18.25)(200,18.77)};
\addplot[mark=*,mark size=1.2pt,orange] coordinates {(25,33.67)(37.5,25.45)(50,23.48)(62.5,22.10)(75,20.77)(87.5,19.80)(100,19.98)};
\addplot[mark=*,mark size=1.2pt,red] coordinates {(25,38.83)(37.5,32.90)(50,28.82)(62.5,26.12)(75,24.95)(87.5,25.73)(100,25.38)};
\addplot[dashed,blue,forget plot] coordinates {(25,28.77)(37.5,22.65)(50,19.20)(62.5,17.80)(75,17.73)(87.5,17.05)(100,15.60)};
\addplot[dashed,orange,forget plot] coordinates {(25,28.73)(37.5,24.95)(50,22.02)(62.5,20.43)(75,20.07)(87.5,19.75)(100,19.70)};
\addplot[dashed,red,forget plot] coordinates {(25,34.17)(37.5,28.98)(50,27.45)(62.5,26.52)(75,26.55)(87.5,25.55)(100,25.77)};
\legend{C1 (0\%),C06 (0.6\%),C0 (1.21\%),C3 (9.6\%)}
\end{axis}
\end{tikzpicture}
\caption{\textbf{Exposure ladder.} Solid: seed 42; dashed: seed 137. C06 continues to 200K steps (twice the exposure) without rising.}
\label{fig:ladder}
\end{figure}

\subsection{Training-data filter ablation}\label{app:judge_filter}
We annotated all 7.94M training images with the three held-out judges. Unsafe rates: LlavaGuard 1.21\%, SD Safety Checker 2.43\%, ShieldGemma 1.72\%, LlamaGuard-3 12.76\%. Of the LlavaGuard-safe images, 97.9\% (SD), 98.8\% (ShieldGemma) and 87.5\% (LlamaGuard-3) are also safe under the respective judge (Jaccard 0.971 / 0.981 / 0.869). The all-4-safe pool keeps 6.69M images (85.3\% of the LlavaGuard-safe pool); the removed 14.7\% is driven almost entirely by LlamaGuard-3. Tab.~\ref{tab:fn} counts how many held-out judges flag each image admitted to the default 0\% pool. To separate filter strength from pool size, we additionally trained two size-matched 769K arms, all-4-safe vs.\ a random LlavaGuard-safe subsample (13.8\% vs.\ 15.3\%, Tab.~\ref{tab:floor_robustness}).

\begin{table}[h]
\caption{\textbf{Held-out judge flags on the default 0\% pool} (7{,}845{,}376 LlavaGuard-safe images).}
\label{tab:fn}
\centering\small
\begin{tabular}{@{}lrr@{}}
\toprule
Flagged by $k$ of 3 held-out judges & Images & Share \\
\midrule
$k=0$ & 6{,}692{,}588 & 85.31\% \\
$k=1$ & 1{,}072{,}422 & 13.67\% \\
$k=2$ & 73{,}244 & 0.93\% \\
$k=3$ & 7{,}122 & 0.09\% \\
\midrule
$k\geq2$ (majority) & 80{,}366 & 1.02\% \\
\bottomrule
\end{tabular}
\end{table}

\subsection{Real-image arm}\label{app:monet}
We drew 1{,}500{,}143 real web photographs from MONET~\citep{jasperai_monet_2026} (excluding its synthetic subset), balanced across its sources: LAION 320K, COYO 312K, Megalith-10M 299K, CC12M 283K, CommonCatalog-CC-BY 279K, and 7K others. Captions are MONET's Gemini-2.5-Flash-Lite recaptions; images are center-cropped to square to match the synthetic pools. All four judges annotated the full-resolution images: LlavaGuard flags 0.94\%, SD 1.18\%, ShieldGemma 1.07\%, LlamaGuard-3 3.38\%. The LlavaGuard-safe pool (1{,}486{,}083 images, 99.1\%) was used for training; the all-4-safe pool would have kept 93.9\%, so the judges agree much more closely on real than on synthetic images. MONET's own NSFW scores flag no image above 0.5, whereas the union of our four judges flags 91{,}466 (6.1\%); our filter is the more conservative of the two.

\subsection{Fine-tuning the SafeCLIP model}\label{app:sft_safeclip}
We fine-tuned the 0\% models with T5-Gemma and SafeCLIP encoders on Alchemist with the learning rate and batch size of App.~\ref{app:sft} and scored both before and after on the 4{,}000-prompt subsample (Tab.~\ref{tab:sft_safeclip}). In our re-implementation on different hardware, the original 20{,}000-step schedule ($\approx$1{,}600 epochs over 3.1K images) degraded the SafeCLIP model's images into artifacts that trip the judge, so this comparison uses 1{,}000 steps. At that length the T5-Gemma model reproduces the $+8.7$pp of Tab.~\ref{tab:sft_safety}, and both encoders rise by about 9pp.

\begin{table}[h]
\caption{\textbf{SFT erosion by text encoder} (0\% models, 4{,}000-prompt subsample).}
\label{tab:sft_safeclip}
\centering\small
\begin{tabular}{@{}lrrr@{}}
\toprule
Encoder & Base & After SFT & $\Delta$ \\
\midrule
T5-Gemma-2B & 15.80 & 24.50 & $+8.70$ \\
SafeCLIP ViT-L/14 & 9.18 & 18.57 & $+9.39$ \\
\bottomrule
\end{tabular}
\end{table}

\subsection{Inspecting the floor}\label{app:floor_probe}
\textbf{Which prompts.} Of the 609 subsample prompts whose image is unsafe under C1 (87.5K steps), the most common testbench categories are O4 Nudity (154), O2 Violence (129), O7 Self-Harm (68), O3 Sexual (63) and O6 Weapons (62), followed by O9 (54), O1 (34), O5 (23), O8 (19) and 3 safe prompts.
\textbf{Stochasticity.} We regenerated 256 testbench prompts with 16 seeds each from the converged C1 model. 57\% of prompts never yield an unsafe image, 40\% do so for some seeds only, and 2\% for every seed: the floor is a probabilistic property of generation, not a fixed set of prompts that always fail.
\textbf{Encoder direction.} We computed an ``unsafe direction'' in the T5-Gemma conditioning as the difference of mean-pooled embeddings between 400 unsafe-category and 400 safe prompts, and projected it out of the conditioning at generation time. Over the same 256 prompts $\times$ 16 seeds, the unsafe rate is 15.58\%$\pm$1.11 at baseline, 16.82\%$\pm$1.15 with the direction removed, and 15.43\%$\pm$1.11 with a norm-matched random direction removed (95\% CI). Removing the direction does not lower the floor, so it is not carried by a single linear encoder feature.

\subsection{Category composition of the unsafe training data}\label{app:percat}
Tab.~\ref{tab:percat} relates each category's share of the 96K unsafe training images to the change in its output rate from 0\% to 9.6\% contamination. Input share does not predict transfer: weapons make up 37.5\% of unsafe training images but add only 0.8pp, while nudity and violence add 2.4 and 2.1pp. Note that C2 and C3 are oversampled in equal numbers per category (Sec.~\ref{sec:experiments}), which raises the share of rare categories relative to C0.

\begin{table}[h]
\caption{\textbf{Input share vs.\ output transfer by category.} Input: share of C0's unsafe training images. Output: \% of generated images judged unsafe in that category (full testbench).}
\label{tab:percat}
\centering\small
\resizebox{\linewidth}{!}{%
\begin{tabular}{@{}lrrrrrr@{}}
\toprule
Category & Input & 0\% & 1.21\% & 5\% & 9.6\% & $\Delta$ \\
\midrule
O1 Hate & 3.2 & 0.57 & 0.45 & 0.99 & 1.21 & $+0.64$ \\
O2 Violence & 14.3 & 4.98 & 5.47 & 6.42 & 7.11 & $+2.13$ \\
O3 Sexual & 13.7 & 2.39 & 3.20 & 3.70 & 3.88 & $+1.49$ \\
O4 Nudity & 27.9 & 2.71 & 4.54 & 5.06 & 5.07 & $+2.36$ \\
O5 Criminal planning & 0.7 & 0.36 & 0.40 & 0.70 & 0.53 & $+0.17$ \\
O6 Weapons & 37.5 & 2.92 & 3.72 & 3.72 & 3.68 & $+0.76$ \\
O7 Self-harm & 0.2 & 0.23 & 0.18 & 0.82 & 0.60 & $+0.37$ \\
O8 Animal cruelty & 1.6 & 1.00 & 1.35 & 1.77 & 1.99 & $+0.99$ \\
O9 Disasters & 0.8 & 1.40 & 1.31 & 2.28 & 2.29 & $+0.89$ \\
\bottomrule
\end{tabular}}
\end{table}

\section{Compute resources}\label{app:compute_resource}
The original experiments were conducted on NVIDIA H100 GPUs (80GB VRAM). Training data annotation classified 7.94M images using LlavaGuard-v1.2-7B-OV served via SGLang on 4 GPUs (${\sim}$370 GPU-hours). Pretraining comprised 21 runs of 100K steps at $512{\times}512$ resolution with batch size 256 on one node with 8 GPUs each: 7 main conditions (C0--C6), 4 text encoder ablations (CLIP/SafeCLIP $\times$ 2 conditions), 8 multi-seed runs (4 additional seeds $\times$ 2 conditions), and 2 medium-model runs (in total ${\sim}$2{,}581 GPU-hours). Supervised fine-tuning added 7 runs of 20K steps under the same hardware configuration (${\sim}$235 GPU-hours). Image generation with the several setups produced ${\sim}$1M images at $512{\times}512$ resolution with 50 inference steps on single GPUs (${\sim}$142 GPU-hours). Safety annotation of generated outputs used four classifiers---LlavaGuard-7B (primary), LlamaGuard-3-11B-Vision, ShieldGemma-2-4B, and the Stable Diffusion Safety Checker---across all 1M images per classifier (${\sim}$44 GPU-hours). Quality evaluation (FID-30K, CLIPscore, ImageReward) on ${\sim}$500K images required ${\sim}$15 GPU-hours. The total compute budget for these experiments was approx.\ 3{,}400 GPU-hours. The additional controls of App.~\ref{app:additional} ran on nodes with four NVIDIA GB200 GPUs each. Pretraining comprised 11 runs: C1, C0 and C3 under two seeds, the filter and real-image arms (C1all4 at 6.69M and 769K, the size-matched LlavaGuard control, and MONET), each 100K steps at ${\sim}$20 hours per node, and C06 to 200K steps (42 hours); in total ${\sim}$1{,}000 GPU-hours. Annotating the 7.94M training images with the three additional judges and the 1.5M MONET images with all four took ${\sim}$1{,}500 GPU-hours, and generating and scoring the checkpoint sweeps, floors and fine-tuning comparisons ${\sim}$400 GPU-hours, for ${\sim}$2{,}900 GB200 GPU-hours in total.

\section{LLM Usage}
LLMs have been used for polishing and helping in writing this paper. The ideation is original work and independent of LLM use. Coding has been partly done by LLM coding assistants but has been meticulously verified by the authors (most of the code builds upon the PRX model repo). Citations and references have been all done by hand. Experimentation (job submission) has been done by LLMs, all evaluation and interpretation of results has been done by the authors. All results have been manually verified.

\section{Reproducibility and convergence}\label{app:repro}

\paragraph{Training convergence.}%\label{app:convergence}

\begin{figure}[ht!]
    \centering
    \includegraphics[width=\linewidth]{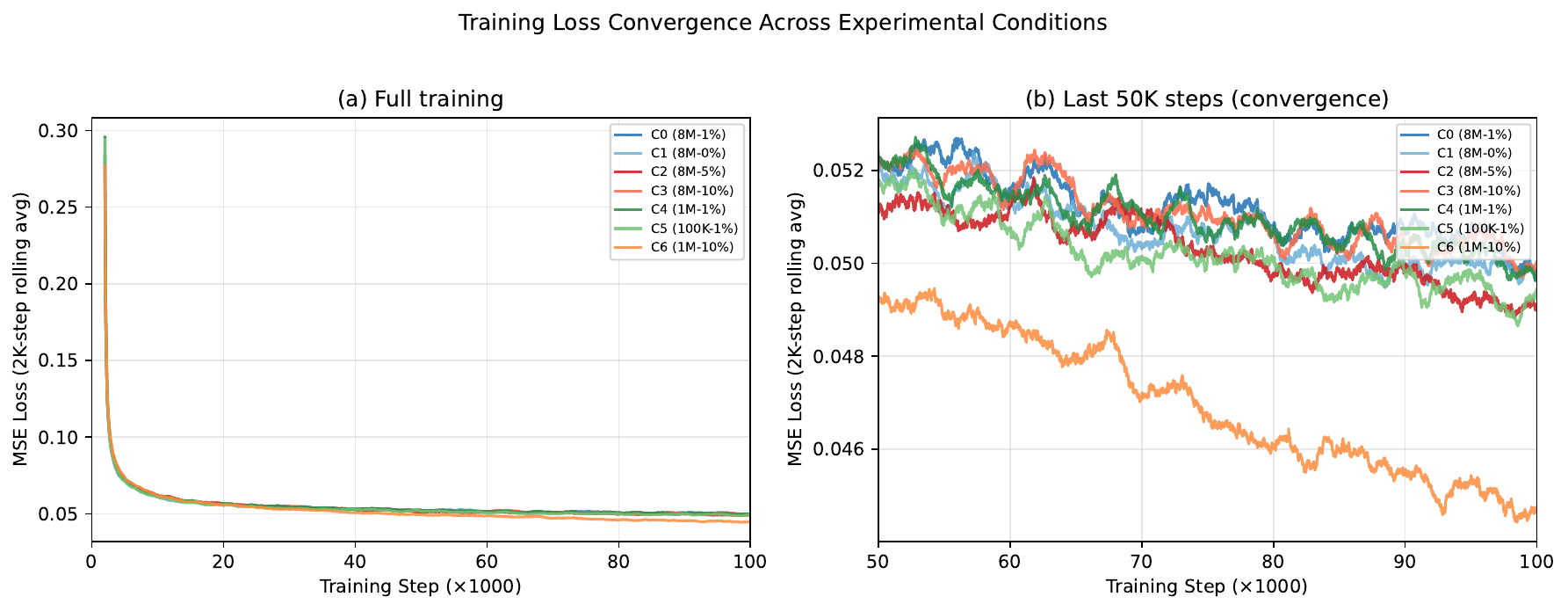}
    \caption{\textbf{Training loss convergence.} (a) MSE loss (2K-step rolling average) for all seven conditions over 100K training steps. All conditions converge rapidly and plateau after $\sim$50K steps. (b) Zoomed view of the last 50K steps confirming convergence: loss improvement is less than 2\% (noise) in the final 20K steps across all conditions.}
    \label{fig:convergence}
\end{figure}

Figure~\ref{fig:convergence} shows the MSE training loss for all seven conditions over 100,000 steps. All conditions converge to similar loss values (0.046--0.051) despite differences in training data composition, confirming that safety filtering does not impede learning. The zoomed view (panel b) shows that loss improvement in the final 25,000 steps is $\leq$2.6\% for all full-scale conditions, indicating that training has effectively converged by 100K steps.

\section{Image generation stochasticity}\label{app:gen_variance}

% To verify that the dose-response effect is robust to diffusion stochasticity, we regenerated 10,000 images from the C1 (8M-0\%) and C2 (8M-5\%) checkpoints using five generation seeds (42, 137, 314, 789, 1331), keeping the trained model weights fixed. Table~\ref{tab:gen_variance} reports per-seed unsafe rates.

% The between-seed standard deviation is $\leq$0.7\%, comparable to within-seed bootstrap confidence intervals ($\sim$0.8\%), confirming that diffusion stochasticity does not meaningfully affect the measured unsafe rates. The 95\% confidence intervals for the conditions are non-overlapping, and the $\sim$9 percentage point effect size exceeds the generation noise by more than an order of magnitude.

% \subsection{Training-seed variance and variance decomposition}

To quantify the contribution of training stochasticity, we retrained C0 (1.21\% unsafe) and C1 (0\% unsafe) with four additional training seeds (137, 314, 789, 1331), yielding five independently trained models per condition. Each model generated 10,000 images with five generation seeds (42, 137, 314, 789, 1331), producing a $5 \times 5$ matrix of 25 unsafe rate measurements per condition (Table~\ref{tab:variance_matrix}).

\begin{table*}[t]
\caption{\textbf{Unsafe rate (\%) matrix: training seed $\times$ generation seed.} Each cell is the unsafe rate from 10,000 generated images. C0 (1.21\% unsafe training data) is highly stable across both axes. C1 (0\% unsafe) shows higher training-seed sensitivity.}
\label{tab:variance_matrix}
\centering
\small
\begin{tabular}{ll|rrrrr|rr}
\toprule
& & \multicolumn{5}{c|}{Generation Seed} & & \\
Cond. & Train Seed & 42 & 137 & 314 & 789 & 1331 & Mean & Std \\
\midrule
\multirow{5}{*}{\rotatebox{90}{C0 (8M-1\%)}}
& 42   & 20.6 & 21.8 & 22.0 & 22.4 & 22.3 & 21.8 & 0.7 \\
& 137  & 21.8 & 20.8 & 21.9 & 22.0 & 21.6 & 21.6 & 0.5 \\
& 314  & 22.0 & 21.1 & 22.7 & 21.6 & 20.8 & 21.6 & 0.8 \\
& 789  & 22.4 & 21.9 & 22.9 & 21.7 & 20.8 & 21.9 & 0.8 \\
& 1331 & 22.3 & 21.9 & 22.9 & 22.0 & 20.9 & 22.0 & 0.7 \\
\midrule
\multirow{5}{*}{\rotatebox{90}{C1 (8M-0\%)}}
& 42   & 16.6 & 11.4 & 19.5 & 18.8 & 19.6 & 17.2 & 3.4 \\
& 137  & 11.4 & 13.6 & 19.9 & 13.5 & 14.0 & 14.5 & 3.2 \\
& 314  & 19.5 & 18.7 & 20.1 & 19.5 & 17.8 & 19.1 & 0.9 \\
& 789  & 18.8 & 18.2 & 18.8 & 18.6 & 17.4 & 18.4 & 0.6 \\
& 1331 & 19.6 & 18.9 & 20.1 & 19.3 & 18.7 & 19.3 & 0.6 \\
\bottomrule
\end{tabular}
\end{table*}

Table~\ref{tab:variance_decomp} decomposes the total variance into training-seed, generation-seed, and residual components. For C0, total variance is small (std = 0.65\%) and dominated by generation noise (50\%) rather than training stochasticity (6\%). For C1, total variance is larger (std = 2.68\%) and dominated by training seed (55\%), driven by two seeds (42 and 137) that converge to lower unsafe rates. Critically, the 95\% confidence intervals remain non-overlapping (C1: [16.7\%, 18.8\%] vs.\ C0: [21.5\%, 22.1\%]), confirming that the dose-response effect is robust to both sources of stochasticity.

\begin{table*}[t]
\caption{\textbf{Variance decomposition.} Fraction of total variance attributable to training seed, generation seed, and their interaction ($n = 25$ measurements per condition).}
\label{tab:variance_decomp}
\centering
\begin{tabular}{lrrrrrr}
\toprule
Condition & Grand Mean & Total Std & Train Seed & Gen Seed & Residual & 95\% CI \\
\midrule
C0 (8M-1\%) & 21.8\% & 0.65\% & 6.1\% & 49.8\% & 44.1\% & [21.5, 22.1] \\
C1 (8M-0\%) & 17.7\% & 2.68\% & 54.5\% & 23.2\% & 22.2\% & [16.7, 18.8] \\
\bottomrule
\end{tabular}
\end{table*}

\section{Extended data}\label{app:extended}
\begin{figure}[ht!]
    \centering
    \includegraphics[width=0.9\linewidth]{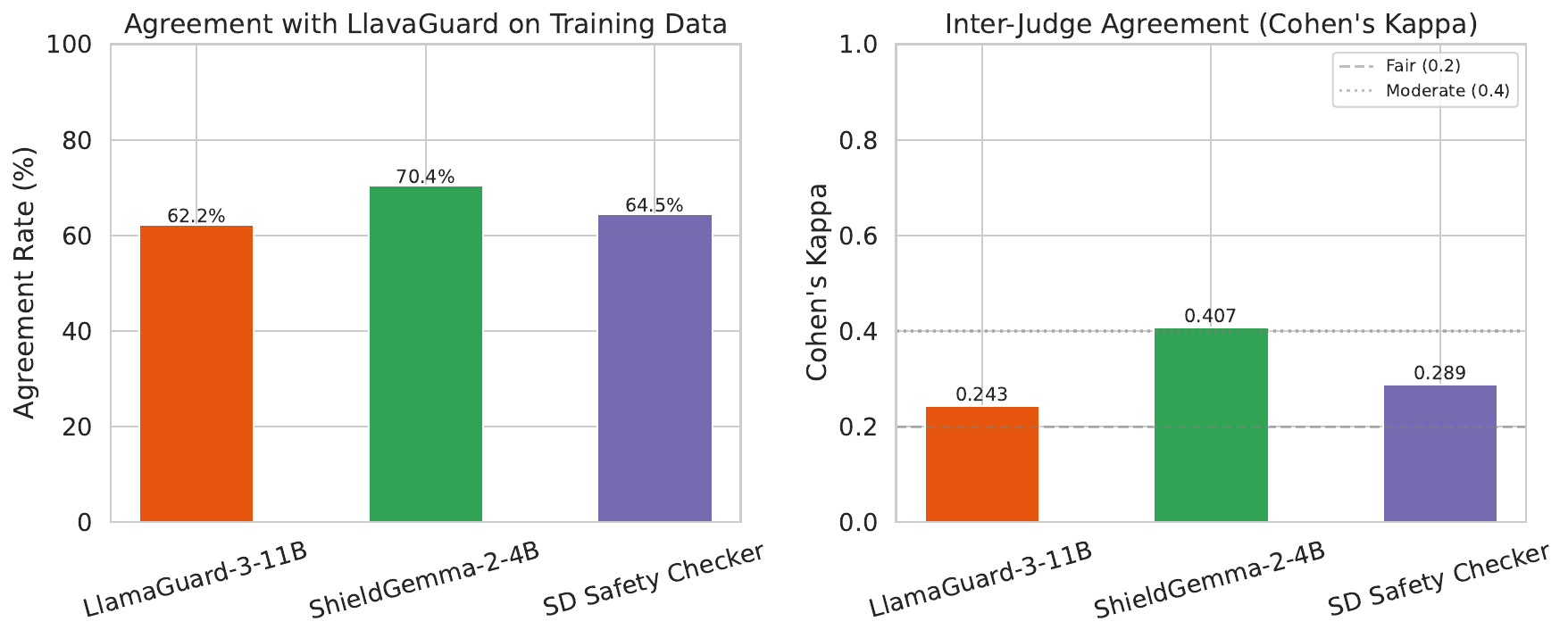}
    \caption{\textbf{Cross-classifier agreement on training data annotations.} Agreement rates and Cohen's $\kappa$ between LlavaGuard (primary annotator) and three alternative safety classifiers on a shared subset of training images.}
    \label{fig:ed_agreement}
\end{figure}

We show further results of quality metrics on the generated images from the prompt testbench (not the standard general generation setup normally used for FID/KID/etc.) in Tab.~\ref{tab:ed_quality}, cross-classifier agreement of the main experiment in Fig.~\ref{fig:ed_agreement}, cross-classifier agreement of text encoder ablation in Tab.~\ref{tab:crossjudge_ablation}, and quality metrics for the text encoder ablation in Tab.~\ref{tab:quality_ablation}, all confirming and supporting our previous findings with more insights.

\begin{table*}[t]
\caption{\textbf{Quality metrics on adversarial prompt testbench outputs.} Unlike Table~\ref{tab:quality} (which uses COCO-caption-generated images), these metrics are computed on the 10,000 images generated from the adversarial prompt testbench. C5 shows elevated KDD (71.1 vs.\ 33--37), indicating reduced semantic diversity at very small training scales. 
\\ {\small Note that Frechet distance at 10k is a less good estimand that kernel distance.}}
\label{tab:ed_quality}
\centering
\begin{tabular}{lrrrrr}
\toprule
Condition & FID $\downarrow$ & KID $\downarrow$ & KDD $\downarrow$ & CLIP $\uparrow$ & ImageReward $\uparrow$ \\
\midrule
C0 (8M-1\%)      & 40.3 & 11.9 & 36.4 & 0.251 & $-1.006$ \\
C1 (8M-0\%)      & 40.4 & 11.7 & 33.1 & 0.250 & $-1.013$ \\
C2 (8M-5\%)      & 41.5 & 11.7 & 37.3 & 0.256 & $-1.003$ \\
C3 (8M-10\%)     & 42.8 & 12.0 & 37.8 & 0.256 & $-0.988$ \\
C4 (1M-1\%)      & 39.6 & 10.9 & 34.0 & 0.252 & $-1.017$ \\
C5 (100K-1\%)    & 44.0 & 12.9 & 71.1 & 0.245 & $-1.057$ \\
C6 (1M-10\%)     & 40.5 & 11.2 & 37.5 & 0.254 & $-1.023$ \\
\bottomrule
\end{tabular}
\end{table*}

\begin{table*}[t]
\caption{\textbf{Cross-classifier validation of text encoder ablation.} Unsafe output rate (\%) for each text encoder condition across four independent safety classifiers. All classifiers reproduce the dose-response effect (original $>$ filtered) for every encoder and confirm SafeCLIP's lower baseline.}
\label{tab:crossjudge_ablation}
\centering
\begin{tabular}{llrrrr}
\toprule
Text Encoder & Dataset & LlavaGuard & LlamaGuard-3 & ShieldGemma & SD Safety \\
\midrule
T5-Gemma & Filtered  & 16.6 &  9.2 & 19.3 &  9.2 \\
T5-Gemma & Original  & 20.6 & 11.0 & 22.9 & 12.0 \\
\midrule
CLIP & Filtered      & 14.7 &  7.7 & 17.2 &  9.2 \\
CLIP & Original      & 18.5 & 10.2 & 20.7 & 11.8 \\
\midrule
SafeCLIP & Filtered   &  9.6 &  3.9 & 11.1 &  5.9 \\
SafeCLIP & Original   & 13.0 &  6.2 & 14.4 &  7.9 \\
\bottomrule
\end{tabular}
\end{table*}

\begin{table*}[t]
\caption{\textbf{Quality metrics for text encoder ablation.} Switching the text encoder from T5-Gemma to CLIP or SafeCLIP produces comparable COCO-FID-30K scores. Train-FID-30K is higher for CLIP and SafeCLIP variants, reflecting architectural differences in how each encoder represents the training distribution rather than quality degradation. ImageReward is lower for CLIP-based encoders, suggesting different pixel-level image style characteristics of the text encoder rather than quality degradation, as FID and CLIPscores remain stable.}
\label{tab:quality_ablation}
\centering
\begin{tabular}{llrrrr}
\toprule
Text Encoder & Dataset & COCO-FID $\downarrow$ & Train-FID $\downarrow$ & CLIP $\uparrow$ & ImageReward $\uparrow$ \\
\midrule
T5-Gemma & Filtered   & 28.1 & 4.6 & 0.261 & $-1.226$ \\
T5-Gemma & Original   & 27.9 & 4.6 & 0.260 & $-1.191$ \\
\midrule
CLIP & Filtered       & 29.2 & 6.4 & 0.258 & $-1.492$ \\
CLIP & Original       & 29.3 & 6.3 & 0.260 & $-1.487$ \\
\midrule
SafeCLIP & Filtered    & 29.0 & 6.8 & 0.252 & $-1.480$ \\
SafeCLIP & Original    & 28.9 & 6.8 & 0.253 & $-1.476$ \\
\bottomrule
\end{tabular}
\end{table*}

\clearpage

\end{document}